\documentclass[10pt,journal,compsoc]{IEEEtran}

\usepackage[nocompress]{cite}
\ifCLASSINFOpdf
\else
\fi
\usepackage{bm}

\usepackage{amsmath}

\usepackage{mdwmath}

\usepackage{eqparbox}

\usepackage{epsfig}
\usepackage{graphicx}
\usepackage{amsmath}
\usepackage{amssymb}
\usepackage{pifont}
\usepackage{url}            
\usepackage{booktabs}       
\usepackage{tabu}           
\usepackage{multirow}
\usepackage{graphicx}
\usepackage{algorithm}
\usepackage{algorithmicx}
\usepackage{algpseudocode}
\usepackage{amsmath,amssymb}
\usepackage[table,dvipsnames]{xcolor}
\usepackage{array}
\usepackage[pagebackref=true,breaklinks=true,colorlinks,bookmarks=false]{hyperref}

\usepackage{cite}

\usepackage{makecell}
\usepackage{tabularx}
\usepackage{multirow}
\usepackage{pifont}
\usepackage{multicol}
\usepackage{array}
\usepackage{adjustbox}

\usepackage{url}            
\usepackage{booktabs}       
\usepackage{amsfonts}       
\usepackage{nicefrac}       
\usepackage{microtype}      
\usepackage{xcolor}         
\usepackage{graphicx}
\usepackage{amsmath}
\usepackage{bm}
\usepackage{multirow}
\usepackage{tabu}
\usepackage{siunitx}
\usepackage{adjustbox}
\usepackage{subcaption}
\usepackage{siunitx}
\usepackage{xcolor}
\usepackage{colortbl}
\usepackage{color}
\usepackage{pifont}
\usepackage{colortbl}
\definecolor{Gray}{gray}{0.95}
\definecolor{Cyan}{rgb}{0.88,1,1}

\usepackage{tabu}           
\usepackage{multirow}
\usepackage{booktabs}
\usepackage{makecell}

\newcommand{\paragrapha}[2][3pt]{\vspace{#1}\noindent\textbf{#2}}

\newcolumntype{x}[1]{>{\centering\arraybackslash}p{#1pt}}
\newlength\savewidth\newcommand\shline{\noalign{\global\savewidth\arrayrulewidth
  \global\arrayrulewidth 1pt}\hline\noalign{\global\arrayrulewidth\savewidth}}

\newcommand{\tablestyle}[2]{\setlength{\tabcolsep}{#1}\renewcommand{\arraystretch}{#2}\centering\footnotesize}

\newcommand{\dynamvit}{DynamicViT}
\newcommand\cb[1]{\color{blue} #1}

\usepackage{soul}
\sethlcolor{Gray}

\def \newpart #1{{#1}}
\definecolor{Grayy}{gray}{0.6}
\newcommand\cg[1]{\color{Grayy} #1}

\begin{document}
%
\title{Dynamic Spatial Sparsification for Efficient Vision Transformers and Convolutional Neural Networks}
%
%
%
%

\author{Yongming Rao$^{*}$,
        Zuyan Liu$^{*}$,
        Wenliang Zhao,
        Jie Zhou, 
        Jiwen Lu$^{\dagger}$
\IEEEcompsocitemizethanks{
\IEEEcompsocthanksitem The authors are with Beijing National Research Center for Information Science and Technology (BNRist), the State Key Lab of Intelligent Technologies and Systems, and the Department of Automation, Tsinghua University, Beijing, 100084, China.
Email:raoyongming95@gmail.com;~liuzuyan19@gmail.com;~zhaowl20@mails.tsinghua.edu.cn;~jzhou@tsinghua.-edu.cn;~lujiwen@tsinghua.edu.cn. 

\IEEEcompsocthanksitem $^{*}$Equal contribution. ~~\textsuperscript{\dag}Corresponding author. 
}
}

%
%

\markboth{IEEE TRANSACTIONS ON PATTERN ANALYSIS AND MACHINE INTELLIGENCE}%
{}
%



\IEEEtitleabstractindextext{%
\begin{abstract}
In this paper, we present a new approach for model acceleration by exploiting spatial sparsity in visual data. 
We observe that the final prediction in vision Transformers is only based on a subset of the most informative regions, which is sufficient for accurate image recognition. Based on this observation, we propose a dynamic token sparsification framework to prune redundant tokens progressively and dynamically based on the input to accelerate vision Transformers. Specifically, we devise a lightweight prediction module to estimate the importance of each token given the current features. The module is added to different layers to prune redundant tokens hierarchically. 
While the framework is inspired by our observation of the sparse attention in vision Transformers, we find that the idea of adaptive and asymmetric computation can be a general solution for accelerating various architectures. We extend our method to hierarchical models including CNNs and hierarchical vision Transformers as well as more complex dense prediction tasks. To handle structured feature maps, we formulate a generic dynamic spatial sparsification framework with progressive sparsification and asymmetric computation for different spatial locations. By applying lightweight fast paths to less informative features and expressive slow paths to important locations, we can maintain the complete structure of feature maps while significantly reducing the overall computations. Extensive experiments on diverse modern architectures and different visual tasks demonstrate the effectiveness of our proposed framework. By hierarchically pruning 66\% of the input tokens, our method greatly reduces 31\% $\sim$ 35\%  FLOPs and improves the throughput by over 40\% while the drop of accuracy is within 0.5\% for various vision Transformers. By introducing asymmetric computation, a similar acceleration can be achieved on modern CNNs and Swin Transformers. Moreover, our method achieves promising results on more complex tasks including semantic segmentation and object detection. Our results clearly demonstrate that dynamic spatial sparsification offers a new and more effective dimension for model acceleration.  Code is available at \url{https://github.com/raoyongming/DynamicViT}.
\end{abstract}

}

\maketitle

\IEEEdisplaynontitleabstractindextext

%
\IEEEpeerreviewmaketitle

\ifCLASSOPTIONcompsoc
\IEEEraisesectionheading{\section{Introduction}\label{sec:introduction}}
\else
\section{Introduction}
\label{sec:introduction}
\fi

\IEEEPARstart{T}{hese} \newpart{years have witnessed the great progress in computer vision brought by the evolution of convolution-based architectures~\cite{he2016deep,krizhevsky2012alex} and recent vision Transformer architectures~\cite{dosovitskiy2020vit,liu2021swin}. As deep neural networks achieve remarkable success on various vision tasks, the models are becoming deeper and wider nowadays. Therefore, exploring the efficient counterparts of these large models becomes an emerging topic that attracts growing attention. Mainstream methods for the acceleration of deep neural networks aim to exploit the sparsity in the channel dimension by pruning the filters that are of less importance~\cite{he2017channel,lin2017runtime,yu2018slimmable} or designing models with fewer filters~\cite{howard2017mobilenets,sandler2018mobilenetv2,wu2019fbnet}.}

\newpart{Informative regions are sparse in images, which motivates many previous methods to study attention mechanism in visual data~\cite{dosovitskiy2020vit,liu2021swin,xu2018attention,cao2017attention}.  A few efforts~\cite{recasens2018learning,wang2020glance} have also been made to explore the spatial sparsity to accelerate model inference by detecting the most informative regions in images and then using the cropped images with fewer pixels to accelerate inference. These methods utilize constrained sparsity by assuming that the most informative pixels are distributed in a rectangular region. To explore more generic sparsity in spatial dimensions, some previous methods introduce the sparse convolution operation to directly process unstructured feature maps~\cite{verelst2020dynamic,choy20194d,graham2017submanifold}. However, this modification on the basic operation may limit their applications and makes it difficult to achieve actual speed-up.}

\begin{figure*}[t]
    \centering
    \includegraphics[width=\textwidth]{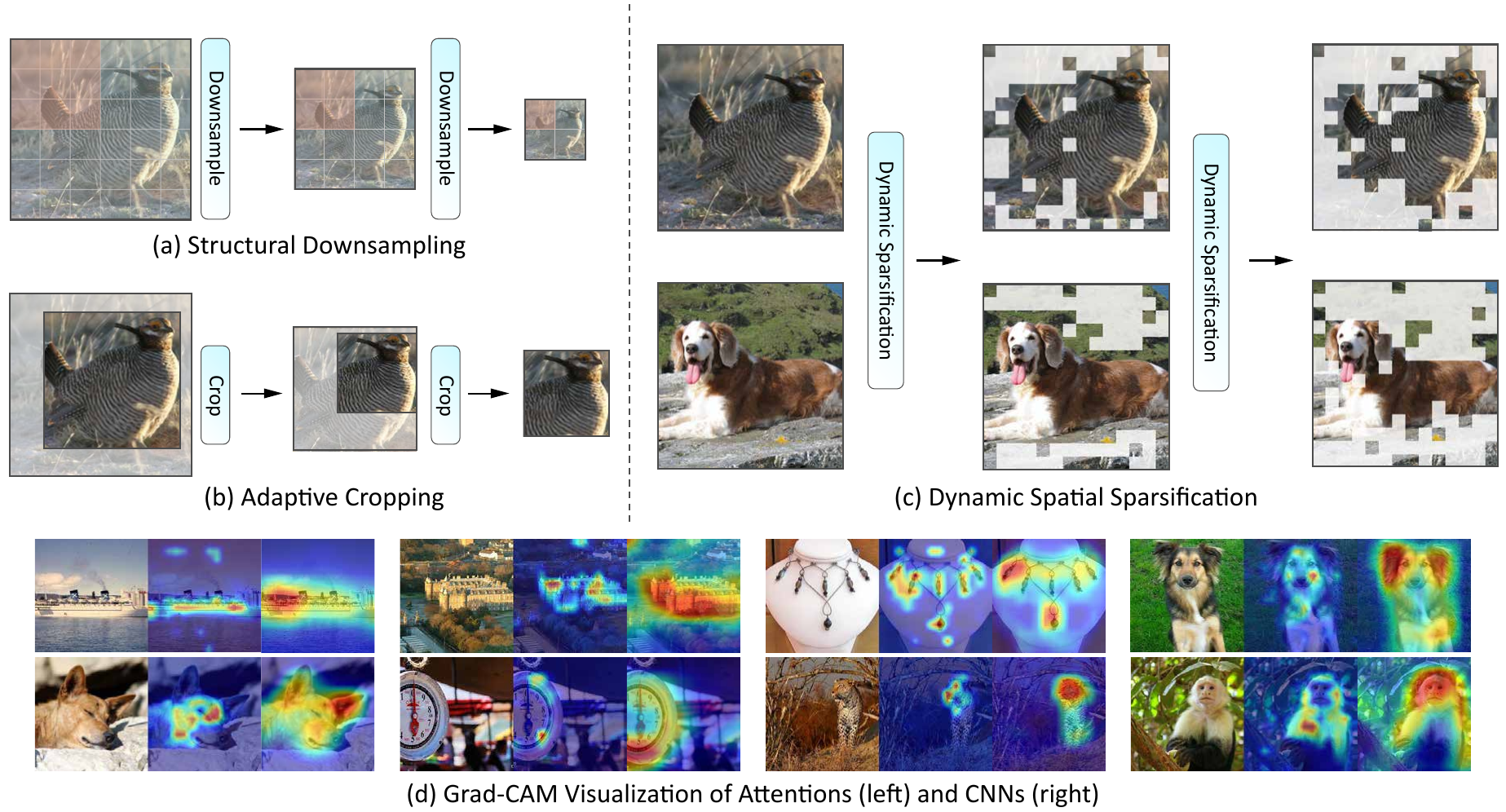}\vspace{-5pt}
    \caption{\textbf{Illustration of our main idea.} CNN models usually leverage the structural downsampling strategy like (a) or cropping the most important rectangular region as (b) to reduce computation related to spatial dimensions. In this paper, we present a \emph{Unstructured} and \emph{data-dependent} dynamic spatial sparsification framework (c) that can better exploit the sparsity in the vision data for efficient acceleration with negligible or no computational cost on less informative regions. (d) visualize the attention and CNN predictions of DeiT-S~\cite{touvron2020deit} and ConvNeXt~\cite{liu2022convnet} model using the visualization method Grad-CAM~\cite{selvaraju2017grad}. Visualization results demonstrate the final prediction in mainstream vision architectures is only based on a subset of the most informative regions, which are distributed non-uniformly in the spatial dimensions. These results suggest it is possible to \emph{selectively} remove a large proportion of \emph{unstructured} regions without hurting the performance.} \vspace{-12pt}
    \label{fig:idea}
\end{figure*}

The emergence of vision Transformers offers us a new way to explore spatial sparsity for learning more efficient models. The way that input is processed by the vision Transformer and its variants, \ie splitting the input image into multiple independent patches, makes it naturally suitable to introduce the spatial sparsity for acceleration. That is, we can prune the tokens of less importance in the input instance, given the fact that many tokens contribute very little to the final prediction (see our visualization in Fig.~\ref{fig:idea}). Since the self-attention module can take the token sequence of variable length as input,  the unstructured sparsified tokens can still be efficiently processed by vision Transformers. Considering that the hierarchical architecture of convolutional neural networks (CNNs) with structural downsampling has improved model efficiency in various vision tasks, we hope to explore the \emph{unstructured} and \emph{data-dependent} downsampling strategy for vision Transformers to further leverage the advantages of self-attention (our experiments also show unstructured sparsification can lead to better performance for vision Transformers compared to structural downsampling). The basic idea of our dynamic spatial sparsification framework is illustrated in Fig.~\ref{fig:idea}.

Based on the dynamic sparsification framework, we first present DynamicViT which learns to make vision Transformers more efficient by dynamically and progressively pruning redundant tokens conditioned on the input. To remove less informative tokens while maximally maintaining the performance, we propose to employ a lightweight prediction module to determine which tokens to be pruned. For each input instance, the prediction module produces a customized binary decision mask to decide which tokens are uninformative. This module is added to multiple layers of the vision Transformer, such that the sparsification can be performed in a hierarchical way as we gradually increase the amount of pruned tokens after each prediction module.
The additional computational overhead introduced by this lightweight module is quite small, especially considering the computational overhead saved by eliminating a large proportion of uninformative tokens. This prediction module can be optimized jointly in an end-to-end manner together with the vision Transformer backbone. To this end, two specialized strategies are adopted. The first one is to adopt Gumbel-Softmax~\cite{eric2017gumbel} to overcome the non-differentiable problem of sampling from a distribution so that it is possible to perform the end-to-end training. The second one is about how to apply this learned binary decision mask to prune the unnecessary tokens. Considering the number of zero elements in the binary decision mask is different for each instance, directly eliminating the uninformative tokens for each input instance during training will make parallel computing unfeasible. Moreover, this would also hinder the back-propagation for the prediction module, which needs to calculate the probability distribution of whether to keep the token even if it is finally eliminated. Besides, directly setting the abandoned tokens as zero vectors is also not a wise idea since zero vectors will still affect the calculation of the attention matrix. Therefore, we propose a strategy called attention masking where we drop the connection from abandoned tokens to all other tokens in the attention matrix based on the binary decision mask. By doing so, we can overcome the difficulties described above. We also modify the original training objective of the vision Transformer by adding a term to constrain the proportion of pruned tokens after a certain layer. During the inference phase, we can directly abandon a fixed amount of tokens after certain layers for each input instance as we no longer need to consider whether the operation is differentiable, and this will greatly accelerate the inference.

While the proposed spatial sparsification framework naturally fits the self-attention and MLP layers in vision Transformers for parallel computing, the strategy of removing a certain subset of the spatial features is not compatible with structure-sensitive operations including downsampling, convolution, and shifted window~\cite{liu2021swin} in CNNs and hierarchical vision Transformers since such operations depend crucially on the neighboring structure of the feature maps. Simply dropping the local features would also influence the well-trained parameters of the filters to a large extent. To address this issue and extend our framework to a wider range of vision architectures, we propose a new approach to represent the uninformative local features. Inspired by previous work like SlowFast~\cite{feichtenhofer2019slowfast} that illustrates the potential of asymmetric networks in temporal modeling, we manage to exploit asymmetry in spatial dimensions. Specifically, similar to our framework for ViT, we adopted the lightweight prediction module to predict the importance of each local feature, dividing the whole image into two parts based on the sparsification ratio determined in advance. The two parts share the structure-sensitive operations for information interchange in the spatial dimensions while having different point-wise layers with asymmetric computational complexity. Since point-wise layers (\ie, convolution layer with a kernel size of 1 or linear layer) contribute the major computation in hierarchical vision Transformers (\eg, 70\% in Swin Transformers~\cite{liu2021swin}) and modern CNNs (\eg, 95\% in ConvNeXt~\cite{liu2022convnet}), applying a group of lightweight point-wise layers (\ie, fast paths) on the less important part will significantly reduce the overall computation while maximally maintain the overall architecture. By applying our new framework to the widely used Swin Transformers and modern convolutional network ConvNeXt, we present the new DynamicSwin and DynamicCNN as the dynamic and more efficient counterparts of these state-of-the-art models.

Extensive experiments on both isotropic vision Transformers and hierarchical models demonstrate the effectiveness of our framework. For isotropic vision Transformers, we illustrate the effectiveness of our method on ImageNet using DeiT~\cite{touvron2020deit} and LV-ViT~\cite{jiang2021token} as the backbones. The experimental results demonstrate the competitive trade-off between speed and accuracy. In particular, by hierarchically pruning 66\% of the input tokens, we can greatly reduce 31\% $\sim$ 35\% GFLOPs and improve the throughput by over 40\% while the drop of accuracy is within 0.5\% for all different vision Transformers.
For hierarchical models, by applying our method to ConvNeXt, we can reduce over 20\% FLOPs without any accuracy drop and over 30\% FLOPs within only a negligible 0.5\% accuracy drop. For Swin Transformers, our method can reduce over 20\% FLOPs without an accuracy drop. Since the new spatial sparsification framework does not affect the structure of the output feature maps, our method is also suitable for downstream tasks like semantic segmentation and object detection. On the widely used semantic segmentation benchmark ADE20k~\cite{zhou2017scene}, we show that our method can reduce $18\%$ of the FLOPs without dropping on mIoU; For object detection on COCO~\cite{lin2014coco}, our method can reduce $15\%$ of the FLOPs with neglectable dropping on AP$^{\rm box}$ and AP$^{\rm mask}$.
Our method demonstrates the possibility of exploiting the sparsity in space for the acceleration of various CNN and vision Transformer models. We expect our attempt to open a new path for future work on the wide-ranging dynamic spatial sparsification and more future applications on downstream tasks. 

This paper is an extended version of our conference paper~\cite{rao2021dynamicvit}. We make several new contributions: 1) We present a more generic dynamic spatial sparsification framework to exploit the sparsity in spatial dimensions for both isotropic vision Transformers and hierarchical models including CNNs and hierarchical vision Transformers; 2) We conducted extensive experiments to evaluate the new framework on representative CNNs and hierarchical vision Transformers. We also show the potential of our framework on more complex dense prediction tasks; 3) We provide more in-depth analyses, ablation studies, and visualizations.

\section{Related Work}
\paragrapha{Vision Transformers. } Transformer model is first widely studied in NLP community~\cite{vaswani2017attention}. Recent progress has demonstrated the variants of Transformers can also be competitive alternatives to CNNs and achieve promising results on different vision tasks including image classification~\cite{dosovitskiy2020vit,touvron2020deit,liu2021swin,zhou2021deepvit}, object detection~\cite{carion2020end}, semantic segmentation~\cite{SETR,cheng2021maskformer} and 3D analysis~\cite{yu2021pointr,zhao2020point}. ViT~\cite{dosovitskiy2020vit} is the first work to directly apply Transformer architecture on non-overlapping image patches for the image classification task, and the whole framework contains no convolution operation. Compared to CNN-type models, ViT can achieve better performance with large-scale pre-training.  DeiT~\cite{touvron2020deit} proposes many training techniques to train the convolution-free Transformer with only ImageNet-1k~\cite{deng2009imagenet}. LV-ViT~\cite{jiang2021token} further improves the performance by introducing a new training objective called token labeling.  Both ViT and its follow-ups split the input image into multiple independent image patches and transform image patches into tokens for further processing. This makes it easy to incorporate the sparsity in space dimension for these Transformer-like models. 

\paragrapha{Model acceleration. }
Model acceleration techniques are important for the deployment of deep models on edge devices. There are many techniques that can be used to accelerate the inference speed of the deep model, including quantization~\cite{gong2014compressing, wang2019haq}, pruning~\cite{he2017channel, rao2018runtime}, low-rank factorization~\cite{yu2017compressing}, knowledge distillation~\cite{hinton2015distilling, liu2020metadistiller} and so on. There are also many methods that aim at accelerating the inference speed of Transformer models. TinyBERT~\cite{jiao2019tinybert} proposes a distillation method to accelerate the inference of Transformers. Star-Transformer~\cite{guo2019star} replaces the fully connected structure with a star-shaped topology. HVT~\cite{pan2021scalable} proposes hierarchical pooling through merging image patches. Different from conventional acceleration methods for vision models like filter pruning~\cite{he2017channel}, our method aims to introduce a new dimension for model acceleration by exploring spatial sparsity. Concurrent work~\cite{pan2021ia,tang2021patch,fayyaz2022adaptive,marin2021token} of our conference version~\cite{rao2021dynamicvit} and follow-ups~\cite{yin2021adavit,meng2021adavit,xu2021evo,liang2022not,yin2022vit} focus on removing redundant tokens for vision Transformer. For example, Token Pooling~\cite{marin2021token} solves the issue through cost-efficient clustering. EViT~\cite{liang2022not} fuses inattentive tokens to expedite computations. A-ViT~\cite{yin2022vit} computes halting scores to recognize discarded tokens. Compared to the works above, our extension aims to develop a more generic framework for diverse vision architectures and tasks. 

\newpart{\paragrapha{Dynamic Neural Networks. } By introducing input-adaptive weights or structures, dynamic neural networks usually can achieve a better trade-off between complexity and accuracy compared to static models. Previous work learns efficient models by dynamically selecting a subset of channels~\cite{yu2018slimmable,lin2017runtime}, resampling input data~\cite{yue2021vision}, skipping unimportant layers~\cite{wang2018skipnet}, or finding an optimal path inside the network~\cite{rao2018runtime,li2020learning,liu2018dynamic}. Some methods accelerate inference by finding the most important region in an image~\cite{wang2020glance,recasens2018learning,najibi2019autofocus} or frames in a video~\cite{wu2019adaframe}. But a generic dynamic network to explore sparsity in spatial dimensions for model acceleration has been barely visited in previous work. In this paper, we show that the dynamic network is a natural and promising way to explore spatial sparsity to learn more efficient vision models. }

\section{Method}

In this section, we will present the details of our dynamic sparsification framework for different types of vision models. We start by elaborating on the detailed designs of our dynamic token sparsification method for vision Transformers and introduce our \dynamvit{} models in Section~\ref{sec:dyvit}. Then, we present DynamicCNN and DynamicSwin models that combine the more generic dynamic spatial sparsification framework and hierarchical architectures in Section~\ref{sec:dycnn}. Lastly, we provide the training and inference details of our method in Section~\ref{sec:train}.

\subsection{Dynamic Sparsification for Vision Transformers}
\label{sec:dyvit}

\begin{figure}[t]
    \centering
    \includegraphics[width=\linewidth]{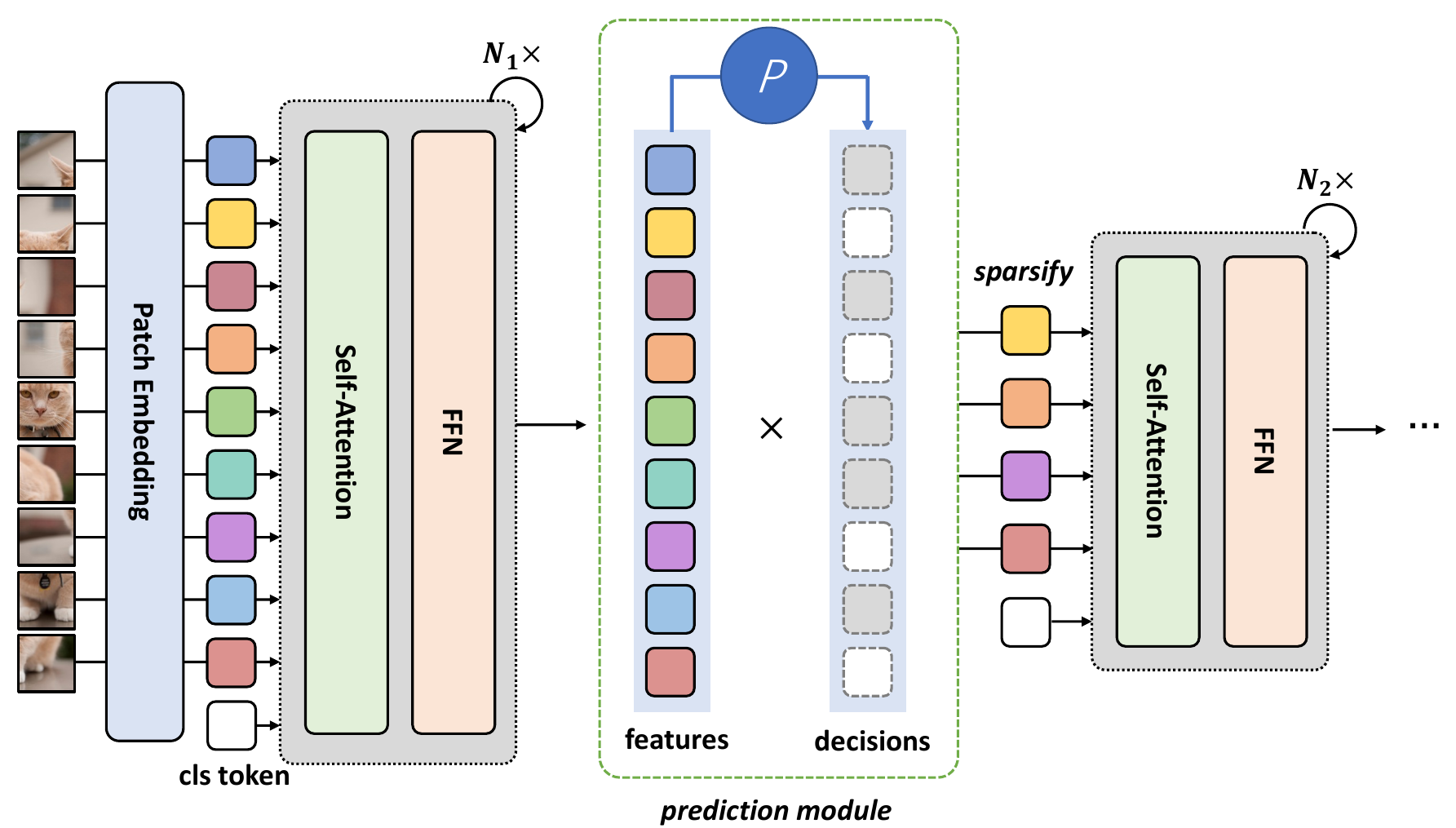}
    \caption{ \textbf{The overall framework of dynamic token sparsification for vision Transformers.} The proposed prediction module is inserted between the Transformer blocks to selectively prune less informative tokens conditioned on features produced by the previous layer. By doing so, fewer tokens are processed in the following layers. }
    \label{fig:overall}
\end{figure}

The overall framework of dynamic spatial sparsification for vision Transformers is illustrated in Fig.~\ref{fig:overall}. Our \dynamvit{} consists of a normal vision Transformer as the backbone and several prediction modules. The backbone network can be implemented as a wide range of vision Transformer (\eg, ViT~\cite{dosovitskiy2020vit}, DeiT~\cite{touvron2020deit}, LV-ViT~\cite{jiang2021token}). The prediction modules are responsible for generating the probabilities of dropping/keeping the tokens. The token sparsification is performed hierarchically through the whole network at certain locations. For example, given a 12-layer Transformer, we can conduct token sparsification before the 4th, 7th, and 10th blocks. During training, the prediction modules and the backbone network can be optimized in an end-to-end manner thanks to our newly devised attention masking strategy. During inference, we only need to select the most informative tokens according to a predefined keeping ratio and the scores computed by the prediction modules. 

\subsubsection{Token Sparsification with Prediction Modules}
An important characteristic of our \dynamvit{} is that the token sparsification is performed hierarchically, \ie, we gradually drop the uninformative tokens as the computation proceeds. To achieve this, we maintain a binary decision mask $\hat{\mathbf{D}}\in \{0, 1\}^{N}$ to indicate whether to drop or keep each token, where $N=HW$ is the number of patch embeddings\footnote{We omit the class token for simplicity, while in practice we always keep the class token (\ie, the decision for the class token is always ``1'').}. We initialize all elements in the decision mask to 1 and update the mask progressively. The prediction modules take the current decision $\hat{\mathbf{D}}$ and the tokens $\bm{x}\in\mathbb{R}^{N\times C}$ as input. We first project the tokens using an MLP:
\begin{equation}
    \bm{z}^\text{local} = \text{MLP}(\bm{x})\in \mathbb{R}^{N\times C'},
\end{equation}
where $C'$ can be a smaller dimension and we use $C'=C/2$ in our implementation. Similarly, we can compute a global feature by:
\begin{equation}
    \bm{z}^\text{global}=\mathcal{A}(\text{MLP}(\bm{x}), \hat{\mathbf{D}})\in \mathbb{R}^{C'},
\end{equation}
where $\mathcal{A}$ is the function that aggregates the information of all the existing tokens and can be simply implemented as an average pooling:
\begin{equation}
    \mathcal{A}(\bm{u}, \hat{\mathbf{D}}) = \frac{\sum_{i=1}^{N}\hat{\mathbf{D}}_i \bm{u}_i}{\sum_{i=1}^{N}\hat{\mathbf{D}}_i}, \quad \bm{u}\in \mathbb{R}^{N\times C'}.
\end{equation}
The local feature encodes the information of a certain token while the global feature contains the context of the whole image, thus both of them are useful. Therefore, we combine the local and global features to obtain local-global embeddings and feed them to another MLP to predict the probabilities to dropping/keeping the tokens:
\begin{align}
    &\bm{z}_i = [\bm{z}_i^\text{local}, \bm{z}^\text{global}],\quad 1\le i\le N,\\
    &\bm{\pi} = \text{Softmax}(\text{MLP}(\bm{z})) \in \mathbb{R}^{N\times 2},
\end{align}
where $\bm{\pi}_{i, 0}$ denotes the probability of dropping the $i$-th token and $\bm{\pi}_{i, 1}$ is the probability of keeping it. We can then generate current decision $\mathbf{D}$ by sampling from $\bm{\pi}$ and update $\hat{\mathbf{D}}$ by
\begin{equation}\label{eq:d}
    \hat{\mathbf{D}}\leftarrow \hat{\mathbf{D}} \odot \mathbf{D},
\end{equation}
where $\odot$ is the Hadamard product, indicating that once a token is dropped, it will never be used.

\subsubsection{End-to-end Optimization with Attention Masking}
Although our target is to perform token sparsification, we find it non-trivial to implement in practice during training. First, the sampling from $\bm{\pi}$ to get binary decision mask $\mathbf{D}$ is non-differentiable, which impedes the end-to-end training. To overcome this, we  apply the Gumbel-Softmax technique~\cite{eric2017gumbel} to sample from the probabilities $\bm{\pi}$:
\begin{equation}
    \mathbf{D} = \text{Gumbel-Softmax}(\bm{\pi})_{*, 1}\in \{0, 1\}^{N},
\end{equation}
where we use the index ``1'' because $\mathbf{D}$ represents the mask of the \emph{kept} tokens. The output of  Gumbel-Softmax is a
one-hot tensor, of which the expectation equals $\bm{\pi}$ exactly. Meanwhile, Gumbel-Softmax is differentiable thus making it possible for end-to-end training.

The second obstacle comes when we try to prune the tokens during training. The decision mask $\hat{\mathbf{D}}$ is usually unstructured and the masks for different samples contain various numbers of 1's. Therefore, simply discarding the tokens where $\hat{\mathbf{D}}_i=0$ would result in a non-uniform number of tokens for samples within a batch,
which makes it hard to parallelize the computation.
Thus, we must keep the number of tokens unchanged while cutting down the interactions between the pruned tokens and other tokens. We also find that merely zeroing out the tokens to be dropped using the binary mask $\hat{\mathbf{D}}$ is not feasible, because in the calculation of self-attention matrix~\cite{vaswani2017attention}
\begin{equation}
    \mathbf{A} = \text{Softmax}\left(\frac{\mathbf{Q}\mathbf{K}^T}{\sqrt{C}}\right)
\end{equation}
the zeroed tokens will still influence other tokens through the \texttt{Softmax} operation. To this end, we devise a strategy called attention masking which can totally eliminate the effects of the dropped tokens. Specifically, we compute the attention matrix by:
\begin{align}
    &\mathbf{P} = \mathbf{Q}\mathbf{K}^T/\sqrt{C} \in \mathbb{R}^{N\times N},\\
    &\mathbf{G}_{ij} = 
    \begin{cases}
    1,& i=j,\\
    \hat{\mathbf{D}}_j,& i\neq j.
    \end{cases}& 1\le i,j\le N,\label{equ:G_ij}\\
    &\tilde{\mathbf{A}}_{ij} =  \frac{\exp(\mathbf{P}_{ij})\mathbf{G}_{ij}}{\sum_{k=1}^N\exp(\mathbf{P}_{ik})\mathbf{G}_{ik}},& 1\le i,j\le N.\label{equ:attention_masking}
\end{align}
By Eq.~\eqref{equ:G_ij} we construct a graph where $\mathbf{G}_{ij}=1$ means the $j$-th token will contribute to the update of the $i$-th token. Note that we explicitly add a self-loop to each token to improve numerical stability. It is also easy to show the self-loop does not influence the results: if $\hat{\mathbf{D}}_j=0$, the $j$-th token will not contribute to any tokens other than itself. Eq.~\eqref{equ:attention_masking} computes the masked attention matrix $\tilde{\mathbf{A}}$, which is equivalent to the attention matrix calculated by considering only the kept tokens but has a constant shape $N\times N$ during training.

\begin{figure}[t]
    \centering
    \includegraphics[width=\linewidth]{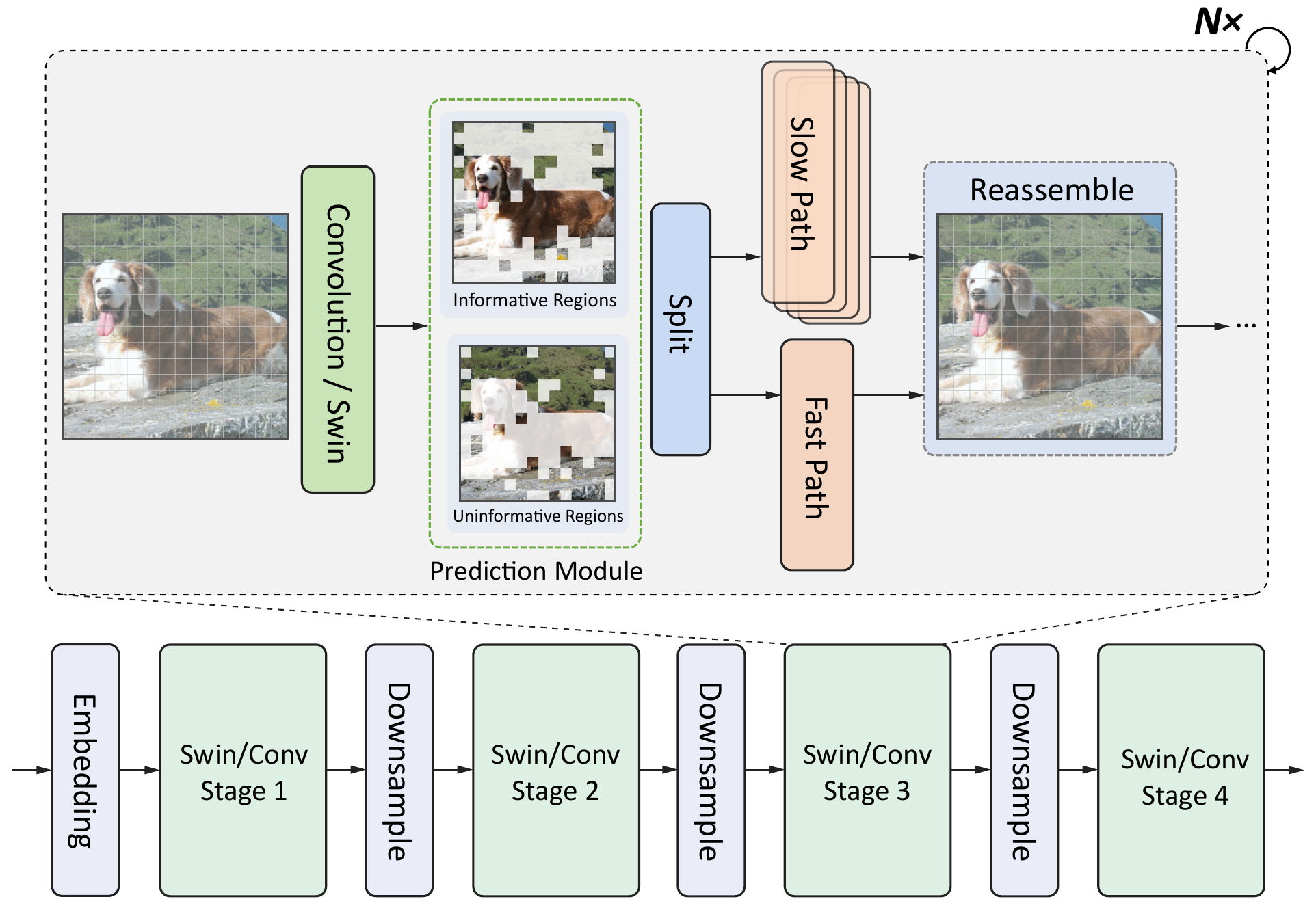}
    \caption{\textbf{The overall framework of asymmetric computation with fast and slow paths for hierarchical models.} After the structure-sensitive operations, the feature maps are split up according to the guidance of the prediction module. Slow path and fast path are applied to informative and uninformative features respectively. Feature maps then are reassembled to maintain the structure for the following layers. The pruning operation is adopted in Stage 3 of hierarchical models which takes up most of the computational cost. } \vspace{-10pt}
    \label{fig:overallnew}
\end{figure}

\subsection{Dynamic Sparsification for Hierarchical Models}\label{sec:dycnn}

The above token sparsification framework that directly removes uninformative tokens naturally fits vision Transformer architecture thanks to the self-attention mechanism that can process variable-length and unstructured features. However, the framework cannot be applied to a wider range of vision models since structure-sensitive operations like convolution and shifted-window attention~\cite{liu2021swin} are the cores of many state-of-the-art hierarchical models including CNNs and hierarchical vision Transformers. To address this issue, we propose a more generic dynamic spatial sparsification framework that introduces asymmetric and dynamic computation to different spatial locations but maintains the structure of the features. The overall framework of dynamic spatial sparsification for hierarchical models is illustrated in Fig.~\ref{fig:overallnew}.

\subsubsection{Asymmetric Computation with Fast and Slow Paths }

Recent state-of-the-art vision models follow the meta-architecture of Transformers~\cite{vaswani2017attention}, where the basic building block consists of a mixing sub-layer like depth-wise convolution layer~\cite{howard2017mobilenets,liu2022convnet} and multi-head self-attention layer~\cite{vaswani2017attention,dosovitskiy2020vit,liu2021swin} to capture spatial relations and an MLP sub-layer (\ie, feed-forward network, FFN) to perform point-wise processing. Given the input feature $\mathbf{x}\in\mathbb{R}^{H\times W\times C}$, the building block can be written as:
\begin{equation}
    \begin{split}
        \bm{x} &\leftarrow\text{Mixer}(\bm{x}), \\
        \bm{x} &\leftarrow \text{FFN}(\bm{x}). \\
    \end{split}
\end{equation}
Since FFNs contribute the majority of computation in such networks and the point-wise layers are also compatible with unstructured features, we propose to introduce asymmetric computation to the FFN sub-layers while keeping the structures of the inputs of the mixing sub-layer unchanged. To this end, we introduce a new lightweight \emph{fast path} $\phi^{\text{fast}}$ to each block and regard the existing FFN layer as the \emph{slow path} $\phi^{\text{slow}}$. Specifically, given the prediction module decision $\mathbf{D}$ representing the importance of each spatial location, we can first \emph{split} the features after the spatial mixing sub-layer into two groups $\bm{x}_1$ and $\bm{x}_2$, which represents the informative and less informative features respectively. Then, we can use the slow and fast paths to process these two groups of features separately. Lastly, the two groups of features are reassembled based on their original spatial positions to ensure that the structure of the output is unchanged. The modified block can be written as:
\begin{equation}
    \begin{split}
        &\bm{x} \leftarrow\text{Mixer}(\bm{x}), \\
        &\bm{x}_1, \bm{x}_2 \leftarrow \text{Split}(\bm{x}, \mathbf{D}), \\
        &\bm{x}_1 \leftarrow \phi^{\text{slow}}(\bm{x}_1),~~ \bm{x}_2 \leftarrow \phi^{\text{fast}}(\bm{x}_2), \\
        &\bm{x} \leftarrow \text{Reassemble}(\bm{x}_1, \bm{x}_2). \\
    \end{split}
\end{equation}
The above framework only assumes that the fast path layer should be a more lightweight replacement for the slow path. We investigate several possible architectural choices for the fast path including linear projection, bottleneck MLP, learnable masking, and zero masking (see our results in Table~\ref{tab:strategy}). Our experiments show that a simple linear projection can achieve a good trade-off between complexity and accuracy and largely outperforms the non-parametric design (\ie, zero mask) and input-independent design (\ie, learnable mask). Therefore, we use linear projection as the fast path in all our models.

\subsubsection{Spatial Sparsification with Prediction Modules }

Our prediction module for vision Transformers that combines local and global features can also be used to predict the importance of each spatial location for hierarchical models. However, due to the architectural difference between isotropic vision Transformers and hierarchical models, we still need to make some modifications to the learning process of the prediction modules: 1) Since we do not remove tokens for hierarchical models, the prediction model can be simply optimized by feature masking. During training, we use both the fast and slow paths to process the input features and combine them based on the predicted binary mask:
\begin{equation}
    \begin{split}
        &\bm{x} \leftarrow\text{Mixer}(\bm{x}), \\
        &\bm{x} \leftarrow \mathbf{D}  \odot  \phi^{\text{slow}}(\bm{x}) +  (1 - \mathbf{D})  \odot  \phi^{\text{fast}}(\bm{x}), \\
    \end{split}
\end{equation}
where the element-wise multiplication is performed by expanding the channel dimension of $\mathbf{D}$. 2) We generate prediction progressively but independently, where we do not add the constraint in Eq.~\ref{eq:d} to update the decision and produce the decision solely based on the current features. Although  decisions for the same spatial location can be opposite at different sparsification levels, we observe our models can generally learn to progressively find uninformative features and exhibit a similar pattern as our models for vision Transformers. 

\paragrapha{Discussion. } Different from previous methods that explore sparsity in vision data by introducing the new sparse convolution operation~\cite{verelst2020dynamic,choy20194d,graham2017submanifold}, we propose a distinct approach that leverages the spatial sparsity by designing a two-path architecture. Our solution has several advantages: 1) Our method can maximally preserve the structure of the feature maps with a progressive sparsification framework. Since the shape of the output of each basic block is not changed after applying our sparsification, the features extracted by our models are still compatible with upsampling and downsampling operations in hierarchical models and downstream applications, which makes our method suitable for a wider range of architectures and tasks. 2)   With limited extra computation, the lightweight fast path sub-layers can mitigate information loss in previous sparsification methods. Some designs~\cite{verelst2020dynamic} in existing sparse ConvNets can be viewed as the zero masking baseline of our method. 3) We do not make assumptions on the types of the spatial mixing sub-layers such that our method is suitable for both modern CNNs and hierarchical vision Transformers, while most previous methods are designed for CNNs.

\subsection{Training and Inference} \label{sec:train}
We now describe the training objectives of our method. The training process includes training the prediction modules such that they can produce favorable decisions and fine-tuning the backbone to make it adapt to spatial sparsification. Assuming we are dealing with a mini-batch of $B$ samples, we adopt the standard cross-entropy loss:
\begin{equation}
    \mathcal{L}_{\rm cls} = \text{CrossEntropy}(\mathbf{y}, \bar{\mathbf{y}}),
\end{equation}
where $\mathbf{y}$ is the prediction of our model (after softmax) and $\bar{\mathbf{y}}$ is the ground truth. 

To minimize the influence on performance caused by our spatial sparsification, we use the original backbone network as a teacher model and hope the behavior of our models is close to the teacher model as possible. Specifically, we consider this constraint from two aspects. First, we make the output features of our models close to the ones of the teacher model, which can be viewed as a kind of self-distillation. For the token sparsification framework for vision Transformers, we supervise the remaining tokens with the corresponding tokens of the teacher models:
\begin{equation}
    \mathcal{L}_\text{distill} = \frac{1}{\sum_{b=1}^B\sum_{i=1}^N \hat{\mathbf{D}}_i^{b, S}}\sum_{b=1}^B\sum_{i=1}^N \hat{\mathbf{D}}_i^{b, S} (\mathbf{t}_{b,i} - \mathbf{t}_{b,i}')^2,
\end{equation}
where $\mathbf{t}_{b,i}$ and $\mathbf{t}_{b,i}'$ denotes the $i$-th token from the $b$-th sample after the last block of our model and the teacher model, respectively. $\hat{\mathbf{D}}^{b, s}$ is the decision mask for the $b$-th sample at the $s$-th sparsification stage. For the spatial sparsification framework, we simply compute the mean squared error between the features of our model and the teacher model:
\begin{equation}
    \mathcal{L}_\text{distill} = \frac{1}{BN}\sum_{b=1}^B\sum_{i=1}^N (\mathbf{t}_{b,i} - \mathbf{t}_{b,i}')^2.
\end{equation}
Second, we minimize the difference in the predictions between our model and its teacher via the KL divergence:
\begin{equation}
    \mathcal{L}_{\rm KL} = \text{KL}\left(\mathbf{y}\| \mathbf{y}'\right),
\end{equation}
where $\mathbf{y}'$ is the prediction of the teacher model. 

Finally, we want to constrain the ratio of the kept tokens to a predefined value. Given a set of target ratios for $S$ stages $\bm{\rho}=[\rho^{(1)}, \ldots, \rho^{(S)}]$, we utilize an MSE loss to supervise the prediction module:
\begin{equation}
    \mathcal{L}_\text{ratio} = \frac{1}{BS}\sum_{b=1}^{B}\sum_{s=1}^S \left(\rho^{(s)} - \frac{1}{N}\sum_{i=1}^N\hat{\mathbf{D}}^{b, s}_i\right)^2.
\end{equation}

The full training objective is a combination of the above objectives:
\begin{equation}
    \mathcal{L} = \mathcal{L}_{\rm cls} + \lambda_{\rm KL}\mathcal{L}_{\rm KL} +  \lambda_{\rm distill}\mathcal{L}_{\rm distill} + \lambda_{\rm ratio}\mathcal{L}_{\rm ratio},
\end{equation}

During inference, given the target ratio $\bm{\rho}$, we can divide the features into two groups based on the probabilities produced by the prediction modules such that only exact $m^s=\lfloor\rho^{(s)}N\rfloor$ features are selected as the informative features at the $s$-th stage. Formally, for the $s$-th stage, let
\begin{equation}
\mathcal{I}^s = \text{argsort}(\bm{\pi}_{*, 1})
\end{equation}
be the indices sorted by the keeping probabilities $\bm{\pi}_{*, 1}$, we can then select the features of which the indices lie in $\mathcal{I}^s_{1:m^s}$ as the informative features while treating the others as the less informative features. For DynamicViT models, we direct discard these less informative features to achieve a higher speed-up ratio. For DynamicCNN and DynamicSwin models, we use the slow path and fast path to process these two groups to preserve the feature structure. In this way, our models introduce spatial sparsity dynamically at runtime, thus can reduce the computational costs during inference.

\section{Experiments}\label{sec:experiemnt}

\newcommand{\imnetacc}{ImageNet Acc. (\%)}
\newcommand{\throughput}{Throughput (im/s)}
\begin{table*}[t]
  \centering
  \caption{\small \textbf{Results of DynamicViT on ImageNet.} We apply our method on two representative vision Transformers: DeiT~\cite{touvron2020deit} and LV-ViT~\cite{jiang2021token} with two different sizes: small and base (medium). DeiT-S and DeiT-B are widely used isotropic vision Transformers with a simple architecture. LV-ViT-S and LV-ViT-M are state-of-the-art vision Transformers. We report the top-1 classification accuracy, theoretical complexity in FLOPs, and throughput for different ratios $\rho$. The throughput is measured on a single NVIDIA RTX 3090 GPU with batch size fixed to 128. }\vspace{-5pt}
  \adjustbox{width=\textwidth}{
    \begin{tabu}to 0.95\textwidth{cl*{4}{X[l]}}\toprule
    \multirow{2}{*}{~~~~~~Base Model~~~~} & \multicolumn{1}{c}{\multirow{2}{*}{Metrics}} & \multicolumn{4}{c}{Keeping Ratio $\rho$ at each stage} \\\cmidrule{3-6}
          &       & 1.0   & 0.9   & 0.8   & 0.7  \\\midrule
    \multirow{3}[0]{*}{DeiT-S~\cite{touvron2020deit}} 
        & ~~~~\imnetacc{}  & 79.8  & 79.8 {\cb(-0.0)}  & 79.6 {\cb(-0.2)}  & 79.3 {\cb(-0.5)} \\
          & ~~~~FLOPs (G) & 4.6   & 4.0 {\cb(-12\%)}   & 3.5  {\cb(-25\%)}   & 3.0  {\cb(-35\%)}  \\
          & ~~~~\throughput{}\qquad \qquad & 1484.3  & 1670.6  {\cb(+13\%)}  & 1962.7  {\cb(+32\%)}  & 2244.3  {\cb(+51\%)}  \\\midrule
    \multirow{3}[0]{*}{DeiT-B~\cite{touvron2020deit}} 
        & ~~~~\imnetacc{}  & 81.8  & 81.8 {\cb(-0.0)}  & 81.6 {\cb(-0.2)}  & 81.4 {\cb(-0.4)} \\
          & ~~~~FLOPs (G) & 17.6   & 15.5 {\cb(-14\%)}   & 13.3 {\cb(-27\%)}   & 11.5  {\cb(-35\%)}  \\
          & ~~~~\throughput{} & 610.9  & 649.9  {\cb(+11\%)}  & 760.3  {\cb(+24\%)}  & 872.9  {\cb(+43\%)}  \\\midrule
    \multirow{3}[0]{*}{LV-ViT-S~\cite{jiang2021token}} 
        & ~~~~\imnetacc{}  & 83.3  & 83.3 {\cb(-0.0)}  & 83.2 {\cb(-0.1)}  & 83.0 {\cb(-0.3)}  \\
          & ~~~~FLOPs (G) & 6.6   & 5.9 {\cb(-11\%)}    & 5.2 {\cb(-21\%)}    & 4.6 {\cb(-31\%)}   \\
          & ~~~~\throughput{} & 1062.0  & 1170.9 {\cb(+10\%)}  & 1341.1 {\cb(+26\%)}   & 1489.9 {\cb(+40\%)}   \\\midrule
    \multirow{3}[0]{*}{LV-ViT-M~\cite{jiang2021token}} 
        & ~~~~\imnetacc{}  & 84.0  & 83.9 {\cb(-0.1)}  & 83.9 {\cb(-0.1)}  & 83.8 {\cb(-0.2)}  \\
          & ~~~~FLOPs (G) & 12.7  & 11.2 {\cb(-12\%)}  & 9.8 {\cb(-23\%)}    & 8.6 {\cb(-32\%)}   \\
          & ~~~~\throughput{} & 636.1  & 727.1 {\cb(+14\%)}   & 838.2 {\cb(+32\%)}  & 952.4 {\cb(+50\%)}  \\\bottomrule
    \end{tabu}%
}
  \label{tab:main}%
\end{table*}%
	
We conduct extensive experiments to evaluate the proposed dynamic spatial sparsification method. We first present the main results on ImageNet~\cite{deng2009imagenet} by applying our framework to representative isotropic vision Transformers, modern CNNs, and hierarchical vision Transformers.  Then, we show the potential of our models for more challenging dense prediction tasks on commonly used semantic segmentation benchmark ADE20k~\cite{zhou2017scene} and object detection dataset COCO~\cite{lin2014coco}. Lastly, we provide visualization results, an in-depth analysis of our framework, and ablation studies on architectural choices and hyper-parameters. 

\subsection{Image Classification}

\subsubsection{Results of DynamicViT} 

\paragrapha{Setups. }\newpart{We start by developing several DynamicViT models by applying our dynamic spatial sparsification method to isotropic vision Transformers DeiT~\cite{touvron2020deit} and LVViT~\cite{jiang2021token}. In our experiments, we fix the number of sparsification stages $S=3$ and apply the target keeping ratio $\rho$ as a geometric sequence $[\rho, \rho^2, \rho^3]$ where $\rho$ ranges from $(0, 1)$. Note that our token sparsification is performed hierarchically in three stages, there are only $\lfloor N\rho^3\rfloor$ tokens left after the last stage. During training DynamicViT models, we follow most of the training techniques used in DeiT~\cite{touvron2020deit}. We use the pre-trained vision Transformer models to initialize the backbone models and jointly train the whole model for 30 epochs. We set the learning rate of the prediction module to $\frac{\text{batch size}}{1024}\times 0.001$ and use $0.01\times$ learning rate for the backbone model. In all experiments of \dynamvit{}, weights of losses are set to $ \lambda_{\rm KL}=0.5,\lambda_{\rm distill}=0.5,\lambda_{\rm ratio}=2$. We fix the weights of the backbone models in the first 5 epochs. All of our models are trained on a single machine with 8 GPUs. More training details can be found in the supplementary material.}

\paragrapha{Main results. } One of the most advantages of the~\dynamvit{} is that it can be applied to a wide range of vision Transformer architectures to reduce the computational complexity with minor loss of performance. In Table~\ref{tab:main}, we summarize the main results on ImageNet where we evaluate our~\dynamvit{} using four base models (DeiT-S~\cite{touvron2020deit}, Deit-B~\cite{touvron2020deit}, LV-ViT-S~\cite{jiang2021token} and LV-ViT-M~\cite{jiang2021token}). We report the top-1 accuracy, FLOPs, and throughput under different keeping ratios $\rho$. We demonstrate that our \dynamvit{} can reduce the computational costs by $31\%\sim 35\%$ and accelerate the inference at runtime by $43\%\sim 51\%$, with the negligible influence on performance ($-0.2\%\sim -0.5\%$).

\paragrapha{Comparisons with the state-of-the-arts. }In Table~\ref{tab:sota}, we compare the~\dynamvit{} with the state-of-the-art models in image classification, including convolutional networks and Transformer-like architectures. We use the \dynamvit{} with LV-ViT~\cite{jiang2021token} as the base model and use ``$/\rho$'' to indicate the keeping ratio. We observe that our \dynamvit{} exhibits favorable complexity/accuracy trade-offs at all three complexity levels. Notably, we find our DynamicViT-LV-M/0.7 beats the EfficientNet-B5~\cite{tan2019efficientnet} and NFNet-F0~\cite{brock2021nfnet}, which are two of the current state-of-the-arts CNN architectures. 

\subsubsection{Results of DynamicCNN and DynamicSwin}

\paragrapha{Setups. } \newpart{For hierarchical models, we apply the sparsification operations at stage 3 which contributes most of the complexity (\eg 27 out of 36 layers in ConvNeXt-Small~\cite{liu2022convnet}). We fix the number of sparsification stages $S=3$ and set keeping ratio $\rho$ as an arithmetic sequence $[\rho, \rho-0.2, \rho-0.4]$ to maintain the similar reduced FLOPs as \dynamvit{}. We perform sparsification to the $[k, 2k, 3k]$-th layer, where $k$ is a hyper-parameter decided by the number of layers in stage 3. We set $k=\lfloor \rm Layer/9\rfloor$ in the experiments. During training our hierarchical sparsification models, we use linear projection as the fast path. Pre-trained weights are used to initialize the backbone and the whole model is jointly trained for 120 epochs as the fast path consists of newly initialized parameters that need more training epochs. We use EMA model~\cite{polyak1992acceleration} to avoid overfitting following~\cite{liu2022convnet} and set $\lambda_{\rm KL}=0.5, \lambda_{\rm distill}=0.5, \lambda_{\rm ratio}=10$. We set the learning rate of the prediction module to $\frac{\rm batch size}{1024}\times 0.001$ and use $0.2\times$ learning rate for the backbone model. Other training techniques are similar to \dynamvit{}. }

\paragrapha{Main results. } \newpart{In Table~\ref{tab:convnext}, we summarize the classification results on ImageNet. We evaluate our sparsification framework on two widely-used hierarchical backbones, ConvNeXt~\cite{liu2022convnet} and Swin-Transformer~\cite{liu2021swin}, and develop the DynamicCNN and DynamicSwin models. Three different sizes (tiny, small, and base) are evaluated for both DynamicCNN and DynamicSwin. We report the top-1 accuracy and FLOPs under different keeping ratios $\rho$ and pruning locations. Our DynamicCNN reduces the computational costs by over $20\%$ with no influence on performance and $34\%$ with a limited influence on performance ($-0.5\%$). Our DynamicSwin reduces the computational costs by over $20\%$ with a neglectable influence on performance. These results demonstrate our dynamic sparsification can also accelerate CNN and Transformer-like hierarchical architectures.}

\begin{table}[t]
  \centering
  \caption{\small \textbf{Comparisons with the state-of-the-arts on ImageNet.} We compare our DynamicViT models with state-of-the-art image classification models with comparable FLOPs and the number of parameters. We use the \dynamvit{} with LV-ViT~\cite{jiang2021token} as the base model and use the ``$/\rho$'' to indicate the keeping ratio. We also include the results of LV-ViT models as references. We divide all models into three groups based on their complexity. The models are sorted according to their top-1 accuracy on ImageNet in each group. Our models are highlighted in \hl{gray}. }
  \vspace{-5pt}
 \setlength{\tabcolsep}{5pt}
    \begin{tabular}{lcccc}\toprule
    \multirow{2}{*}{Model} & Params & FLOPs & Input & Top-1  \\ 
    & (M) & (G) & Resolution & Acc. (\%) \\\midrule
    DeiT-S~\cite{touvron2020deit} & 22.1  & 4.6   & 224   & 79.8  \\
    PVT-Small~\cite{wang2021pvt} & 24.5  & 3.8   & 224   & 79.8  \\
    CoaT Mini~\cite{xu2021coat} & 10.0  & 6.8   & 224   & 80.8  \\
    CrossViT-S~\cite{chen2021crossvit} & 26.7  & 5.6   & 224   & 81.0  \\
    PVT-Medium~\cite{wang2021pvt} & 44.2  & 6.7   & 224   & 81.2  \\
    Swin-T~\cite{liu2021swin} & 29.0  & 4.5   & 224   & 81.3  \\
    T2T-ViT-14~\cite{yuan2021t2t} & 22.0  & 5.2   & 224   & 81.5  \\
    CPVT-Small-GAP~\cite{chu2021cpvt} & 23.0  & 4.6   & 224   & 81.5  \\
    CvT-13~\cite{wu2021cvt} & 20.0 & 4.5 & 224 & 81.6 \\
    PS-ViT/14~\cite{yue2021vision} & 21.3 & 5.4 & 224 & 81.7 \\
    CoaT-Lite Small~\cite{xu2021coat} & 20.0  & 4.0   & 224   & 81.9  \\
    \rowcolor{Gray} DynamicViT-LV-S/0.5 & 26.9  & 3.7   & 224   & 82.0  \\
    ConvNeXt-T~\cite{liu2022convnet} & 29.0 & 4.5 & 224 & 82.1 \\
    \rowcolor{Gray} DynamicViT-LV-S/0.7 & 26.9  & 4.6   & 224   & 83.0  \\\midrule
    RegNetY-8G~\cite{radosavovic2020designing} & 39.0  & 8.0   & 224   & 81.7  \\
    T2T-ViT-19~\cite{yuan2021t2t} & 39.2  & 8.9   & 224   & 81.9  \\
    CvT-21~\cite{wu2021cvt} & 32.0 & 7.1 & 224 & 82.5 \\
    Swin-S~\cite{liu2021swin} & 50.0  & 8.7   & 224   & 83.0  \\
    ConvNeXt-S~\cite{liu2022convnet} & 50.0 & 8.7 & 224 & 83.1 \\
    LV-ViT-S~\cite{jiang2021token} & 26.2 & 6.6 & 224 & 83.3 \\
    EfficientNet-B5~\cite{tan2019efficientnet} & 30.0  & 9.9   & 456   & 83.6  \\
    NFNet-F0~\cite{brock2021nfnet} & 72.0  & 12.4  & 256   & 83.6  \\
    \rowcolor{Gray} DynamicViT-LV-M/0.7 & 57.1  & 8.5   & 224   & 83.8  \\\midrule
    DeiT-Base/16~\cite{touvron2020deit} & 86.6  & 17.6  & 224   & 81.8  \\
    CrossViT-B~\cite{chen2021crossvit} & 104.7  & 21.2  & 224   & 82.2  \\
    T2T-ViT-24~\cite{yuan2021t2t} & 64.1  & 14.1  & 224   & 82.3  \\
    TNT-B~\cite{han2021transformer} & 66.0  & 14.1  & 224   & 82.8  \\
    RegNetY-16G~\cite{radosavovic2020designing} & 84.0  & 16.0  & 224   & 82.9  \\
    Swin-B~\cite{liu2021swin} & 88.0  & 15.4  & 224   & 83.5  \\
    ConvNeXt-B~\cite{liu2022convnet} & 89.0 & 15.4 & 224 & 83.8 \\
    \rowcolor{Gray} DynamicViT-LV-M/0.8 & 57.1  & 9.6   & 224   & 83.9  \\
    LV-ViT-M~\cite{jiang2021token} & 55.8 & 12.7 & 224 & 84.0 \\\bottomrule
    \end{tabular}%
  \label{tab:sota}%
\end{table}%

\begin{table}[t]
  \centering
  \caption{\small  \newpart{\textbf{Results of DynamicCNN and DynamicSwin on ImageNet.} We apply our method to two representative hierarchical models. ConvNeXt~\cite{liu2022convnet} is a CNN-type model with more modern designs and state-of-the-art performance. Swin Transformer~\cite{liu2021swin} is a widely used hierarchical vision Transformer. We report the top-1 classification accuracy on the validation set of ImageNet as well as the number of FLOPs under different sparsification policies. All of our models are trained and tested with $224\times 224$ images. Our DynamicCNN and DynamicSwin can significantly reduce model complexity with only a slight performance drop.}}
  \vspace{-5pt}
  \adjustbox{width=0.48\textwidth}{
    \begin{tabu}to 0.48\textwidth{lccll}\toprule
    \multirow{2}{*}{Base Model} & Keeping & Pruning & Top-1 & FLOPs\\
    & Ratio & Location & Acc.(\%) & (G) \\\midrule
    \multirow{4}[0]{*}{ConvNeXt-T~\cite{liu2022convnet}} & - & - & 82.1 & 4.5 \\
            & 0.9 & [1,2,3] & 82.1 {\cb(-0.0)} & 3.9 {\cb(-13\%)} \\
            & 0.8 & [1,2,3] & 81.9 {\cb(-0.2)} & 3.7 {\cb(-17\%)} \\
            & 0.7 & [1,2,3] & 81.6 {\cb(-0.5)} & 3.6 {\cb(-20\%)} \\\midrule
    \multirow{4}[0]{*}{ConvNeXt-S~\cite{liu2022convnet}} & - & - & 83.1 & 8.7 \\
            & 0.9 & [3,6,9] & 83.1 {\cb(-0.0)} & 6.8 {\cb(-22\%)} \\
            & 0.8 & [3,6,9] & 82.8 {\cb(-0.3)} & 6.3  {\cb(-28\%)} \\
            & 0.7 & [3,6,9] & 82.5 {\cb(-0.6)} & 5.8  {\cb(-34\%)} \\\midrule
    \multirow{4}[0]{*}{ConvNeXt-B~\cite{liu2022convnet}} & - & - & 83.8 & 15.4 \\
            & 0.9 & [3,6,9] & 84.0 {\cb(+0.2)} & 11.9 {\cb(-22\%)} \\
            & 0.8 & [3,6,9] & 83.6 {\cb(-0.2)} & 11.0  {\cb(-28\%)} \\
            & 0.7 & [3,6,9] & 83.4 {\cb(-0.4)} & 10.2  {\cb(-34\%)} \\\midrule \midrule
    \multirow{4}[0]{*}{Swin-T~\cite{liu2021swin}} & - & - & 81.2 & 4.5 \\
            & 0.9 & [1,2,3] & 81.0 {\cb(-0.2)} & 4.2 {\cb(-7\%)} \\
            & 0.8 & [1,2,3] & 81.0 {\cb(-0.2)} & 4.1 {\cb(-9\%)} \\
            & 0.7 & [1,2,3] & 80.9 {\cb(-0.3)} & 4.0 {\cb(-11\%)} \\\midrule
    \multirow{4}[0]{*}{Swin-S~\cite{liu2021swin}} & - & - & 83.2 & 8.7 \\
            & 0.9 & [2,4,6] & 83.4 {\cb(+0.2)} & 7.5 {\cb(-13\%)} \\
            & 0.8 & [2,4,6] & 83.3 {\cb(+0.1)} & 7.2  {\cb(-17\%)} \\
            & 0.7 & [2,4,6] & 83.2 {\cb(+0.0)} & 6.9  {\cb(-21\%)} \\\midrule
    \multirow{4}[0]{*}{Swin-B~\cite{liu2021swin}} & - & - & 83.5 & 15.4 \\
            & 0.9 & [2,4,6] & 83.6 {\cb(+0.1)} & 13.3 {\cb(-13\%)} \\
            & 0.8 & [2,4,6] & 83.4 {\cb(-0.1)} & 12.7  {\cb(-17\%)} \\
            & 0.7 & [2,4,6] & 83.4 {\cb(-0.1)} & 12.1  {\cb(-21\%)} \\\bottomrule
    \end{tabu}%
}
  \label{tab:convnext}%
\end{table}%

\vspace{-5pt}
\subsection{Dense Prediction}

To show the generality of our dynamic spatial sparsification framework, we conduct experiments on the more complex dense prediction tasks including semantic segmentation and object detection. 

\subsubsection{Semantic Segmentation}

\newpart{We first evaluate our dynamic spatial sparsification method on ADE20k~\cite{zhou2017scene}, which is a commonly used semantic segmentation dataset. We follow experiment settings in ConvNeXt~\cite{liu2022convnet} and use light-weight Semantic FPN~\cite{kirillov2019panoptic} framework to show the potential of our dynamic sparsification network for dense prediction tasks. We use DynamicCNN as the backbone and load pre-trained weights from models with keeping ratios $\rho=[0.9, 0.7, 0.5]$. To make our DynamicCNN more effective, we keep the ratio loss $\mathcal{L}_{\rm ratio}$ and distill loss $\mathcal{L}_{\rm distill}$ following the practice of the classification task to use the original ConvNeXt model without pruning to guide our model. We train our model for 40k iterations with a batch size of 32 on the training set and report the mIoU on the validation set following common practice. The results are presented in Table \ref{tab:segmentation}. We observe that by reducing over $22\%$  complexity of the backbone, we can effectively accelerate the semantic segmentation models by $15\%\sim 18\%$ with no drop in performance. }

\subsubsection{Object Detection}

\newpart{ We further evaluate our dynamic spatial sparsification method on object detection and instance segmentation. We conduct our experiment on COCO 2017 dataset~\cite{lin2014coco} with the widely used Mask-RCNN~\cite{he2017mask}  framework. We use DynamicCNN as the backbone and follow the training technique on semantic segmentation. We train our model for 12 epochs using AdamW and set the batch size as 16. The results are presented in Table \ref{tab:detection}. We observe that our DynamicCNN model is applicative for the challenging object detection and instance segmentation tasks, with only a negligible influence on the final performance after reducing 15\% of the FLOPs. The results also suggest our method generalizes well to much larger images considered in the detection task. }

\begin{table}[t]
  \centering
  \caption{\newpart{\small \textbf{Results on ADE20k semantic segmentation} using Semantic FPN~\cite{kirillov2019panoptic} framework. We train our models for 40k iterations with a batch size of 32 to test the effectiveness of our model. We report FLOPs tested with $1024\times 1024$ input. We provide both the overall FLOPs and the backbone FLOPs of the models. We report both single-scale (S.S.) and multi-scale (M.S.) mIoU on the validation set. }}
  \vspace{-5pt}
  \adjustbox{width=0.48\textwidth}{
    \begin{tabu}to 0.48\textwidth{lllll}\toprule
    \multirow{2}{*}{Backbone} & \multicolumn{2}{c}{FLOPs (G)} & \multicolumn{2}{c}{mIoU}\\
    \cmidrule(lr){2-3}  \cmidrule(lr){4-5} & Overall & Backbone & S.S. & M.S. \\
    \midrule
    ConvNeXt-S~\cite{liu2022convnet} & 268 & 182 & 43.3 & 44.1 \\
    DynamicCNN-S & 228\cb{(-15\%)} & 141\cb{(-22\%)} & 43.3\cb{(-0.0)} & 44.4\cb{(+0.3)}\\\midrule
    ConvNeXt-B~\cite{liu2022convnet} & 409 & 321 & 44.0 & 45.1\\
    DynamicCNN-B & 337\cb{(-18\%)} & 249\cb{(-22\%)} & 44.0\cb{(-0.0)} & 45.3\cb{(+0.2)}\\\bottomrule
    \end{tabu}%
}
  \label{tab:segmentation}%
\end{table}%

\begin{table}[t]
  \centering
  \caption{\newpart{\small \textbf{Results on COCO object detection and instance segmentation} using Mask-RCNN~\cite{he2017mask}. We train our models for 12 epochs with a batch size of 16. FLOPs are computed on $1280\times 800$ image following common practice, and both overall FLOPs and backbone FLOPs are reported. We report the mean bounding box AP and mask AP on the validation set. }}
  \vspace{-5pt}
  \adjustbox{width=0.48\textwidth}{
    \begin{tabu}to 0.48\textwidth{lllll}\toprule
    \multirow{2}{*}{Backbone} & \multicolumn{2}{c}{FLOPs (G)} & \multicolumn{2}{c}{mAP}\\
    \cmidrule(lr){2-3}  \cmidrule(lr){4-5} & Overall & Backbone & AP$^{\rm box}$ & AP$^{\rm mask}$ \\\midrule
    ConvNeXt-S~\cite{liu2022convnet} & 348 & 177 & 45.8 & 41.3 \\
    DynamicCNN-S & 309\cb{(-11\%)} & 138\cb{(-22\%)} & 45.6\cb{(-0.2)} & 40.9\cb{(-0.4)} \\\midrule
    ConvNeXt-B~\cite{liu2022convnet} & 486 & 313 & 46.6 & 42.1 \\
    DynamicCNN-B & 415\cb{(-15\%)} & 244\cb{(-22\%)} & 46.4\cb{(-0.2)} & 41.8\cb{(-0.3)} \\\bottomrule
    \end{tabu}%
} \vspace{-10pt}
  \label{tab:detection}%
\end{table}%

\begin{figure}[t]
    \centering
    \includegraphics[width=0.48\textwidth]{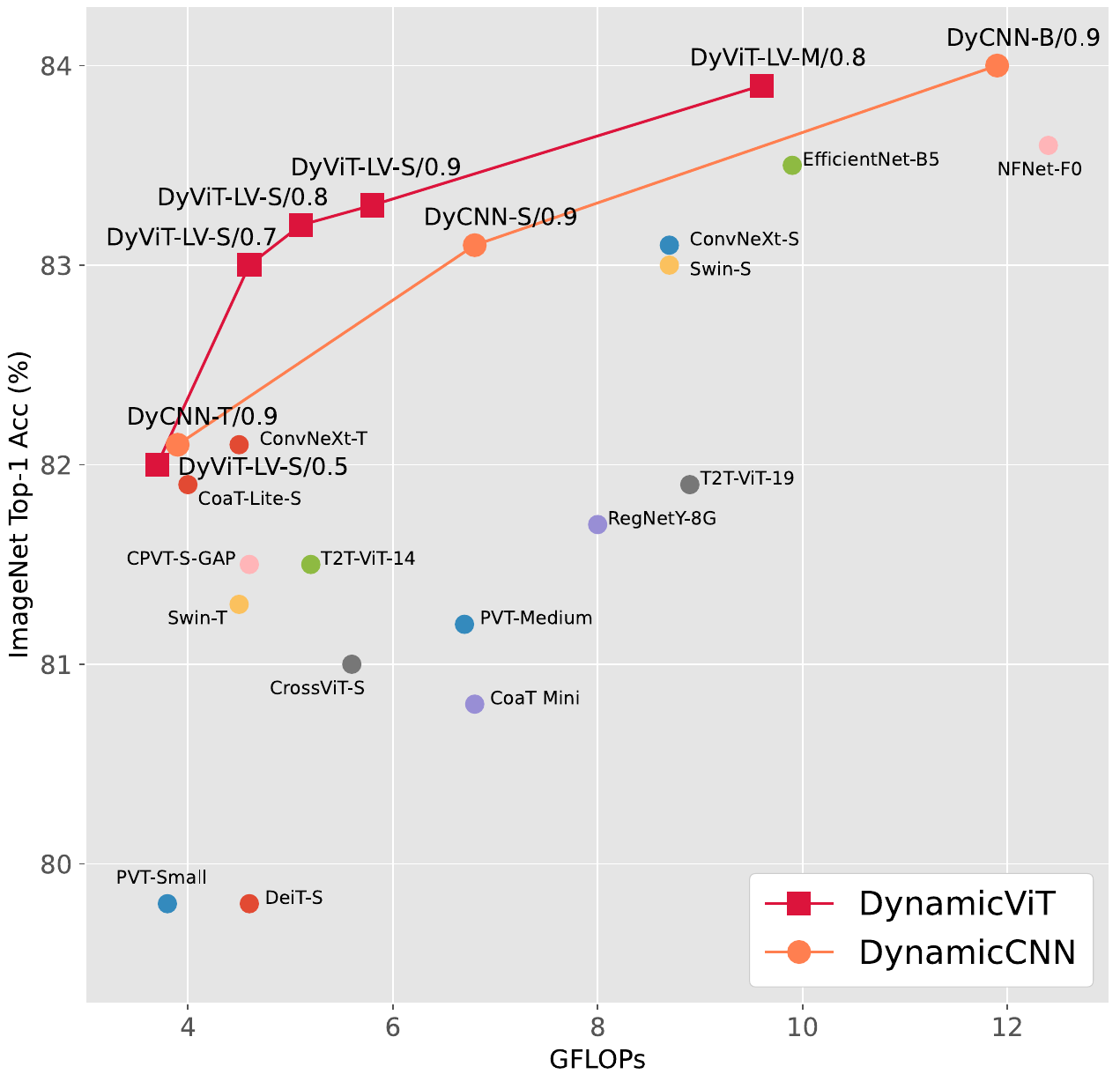}
    \vspace{-5pt}
    \caption{\small\newpart{\textbf{Complexity/accuracy tradeoffs. } We illustrate model complexity (FLOPs) and ImageNet top-1 accuracy trade-offs. We compare DynamicViT and DynamicCNN with the state-of-the-art image classification models. Our models achieve better trade-offs compared to various models. }}
    \label{fig:sota} \vspace{-5pt}
\end{figure}

\subsection{Analysis}

\begin{figure*}[t]
    \centering
    \includegraphics[width=\textwidth]{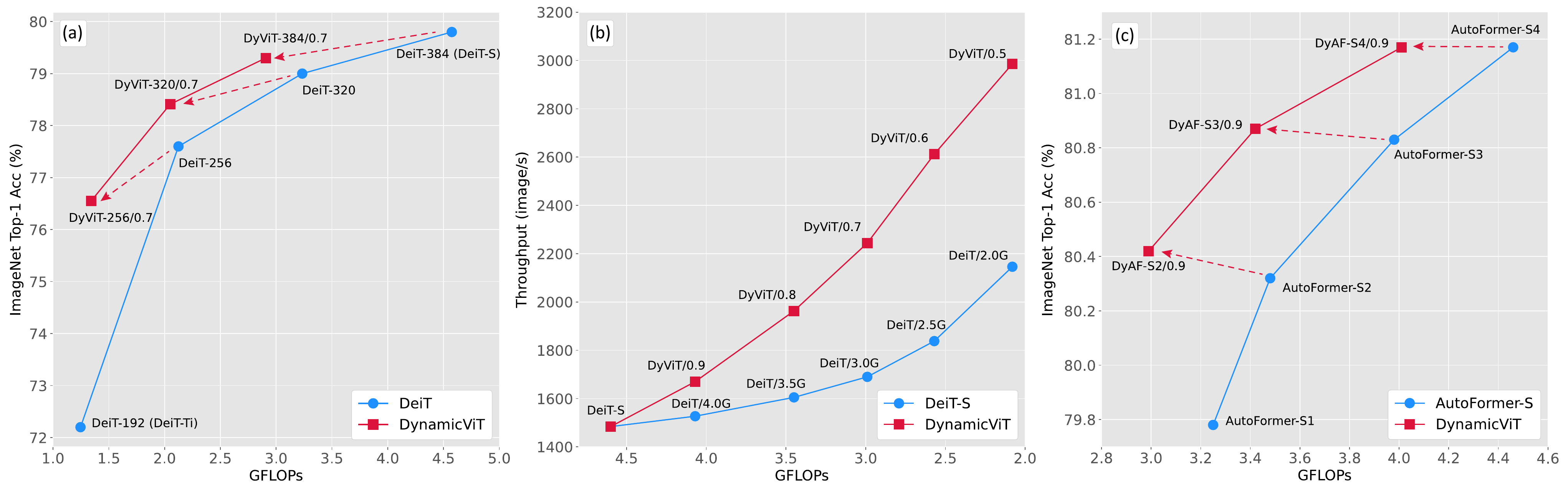}
    \caption{\small\newpart{\textbf{Comparisons with conventional acceleration methods. }In Fig. (a) and Fig. (b), we compare our dynamic token sparsification method with model width scaling. In Fig. (a), We train our \dynamvit{} based on DeiT models with embedding dimensions varying from 192 to 384 and fix ratio $\rho=0.7$. We see dynamic token sparsification is more efficient than commonly used model width scaling. In Fig. (b), we measure the throughput of our \dynamvit{} based on the DeiT-S model and adjust the ratio $\rho$ varying from 0.5 to 1.0. We can clearly observe that \dynamvit{} can achieve a better trade-off between complexity and accuracy and narrow the gap between theoretical complexity and actual throughput on GPU. In Fig. (c), we compare our \dynamvit{} with the conventional acceleration method by applying our method to the state-of-the-art AutoFormer~\cite{chen2021AutoFormer} models. During training, we fix ratio $\rho=0.9$ to fairly compare with AutoFormer. We see our method further improves the trade-off, which suggests spatial sparsity is a new and effective dimension for model acceleration. } }
    \label{fig:tradeoff} \vspace{-5pt}
\end{figure*}

\begin{table*}
\caption{\small \textbf{Comparisons of different sparsification strategies.} We investigate different methods to select redundant tokens based on the DeiT-S model. We report the top-1 accuracy on ImageNet for different methods. We fix the complexity of the accelerated models to 2.9G FLOPs for fair comparisons. The deigns used in our final models are highlighted in \hl{gray}.} 
\centering
\subfloat[\footnotesize Dynamic spatial sparsification \vs static sparsification and structural downsampling.]
{\makebox[0.3\linewidth][c]{
\tablestyle{12pt}{1.2}
\begin{tabular}{c|c}
\textbf{Sparsification Method} & \textbf{Acc. (\%)} \\
\shline
 Structural & 78.2\cb{(-1.6)} \\
Static & 73.4\cb{(-6.4)} \\
\rowcolor{Gray} Dynamic & 79.3(\cb{-0.5}) \\ 
\end{tabular}
\label{Tab:ablation:a}
} 
}
\hfill
\subfloat[\footnotesize Ablation study on different redundant token removal methods.]
{\makebox[0.3\linewidth][c]{
\tablestyle{12pt}{1.2}
\begin{tabular}{c|c}
\textbf{Removal Method} & \textbf{Acc. (\%)} \\
\shline
Random & 77.5\cb{(-2.3)} \\
Attention & 78.1\cb{(-1.7)} \\
\rowcolor{Gray} Prediction & 79.3\cb{(-0.5)} \\ 
\end{tabular}
\label{Tab:ablation:b}
}
}
\hfill
\subfloat[\footnotesize Ablation study on the number of sparsification stages.]
{\makebox[0.3\linewidth][c]{
\tablestyle{12pt}{1.2}
\begin{tabular}{c|c}
\textbf{Number of Stages} & \textbf{Acc. (\%)} \\
\shline
Single-stage & 77.4\cb{(-2.4)} \\
Two-stage & 79.2\cb{(-0.6)} \\
\rowcolor{Gray} Three-stage  & 79.3\cb{(-0.5)} \\ 
\end{tabular}
\label{Tab:ablation:c}
} 
}
\vspace{-5pt}
\label{tab:ablation}
\end{table*}

\newpart{\paragrapha{Complexity/accuracy trade-offs. } We plot the FLOPs-accuracy curve of \dynamvit{} and DynamicCNN series  (where we use DyViT and DyCNN for short), along with other state-of-the-art models to show the complexity and accuracy trade-offs in Fig.~\ref{fig:sota}. We can clearly observe that the proposed dynamic spatial sparsification method can achieve better trade-offs compared to the various vision Transformers as well as carefully designed CNN models, which strongly demonstrates the effectiveness of our method.}

\newpart{\paragrapha{Dynamic sparsification for model scaling.} The success of EfficientNet~\cite{tan2019efficientnet} shows that we can obtain a model with better complexity/accuracy tradeoffs by scaling the model along different dimensions. While in vision Transformers, the most commonly used method to scale the model is to change the number of channels, our \dynamvit{} provides another powerful tool by performing token sparsification. We analyze this nice property of our method in Fig.~\hyperref[fig:tradeoff]{5(a)}. First, we train several DeiT~\cite{touvron2020deit} models with the embedding dimensions varying from 192 (DeiT-Ti) to 384 (DeiT-S). Second, we train our \dynamvit{} based on those models with the keeping ratio $\rho=0.7$. We find that after performing token sparsification, the complexity of the model is reduced to be similar to its variant with a smaller embedding dimension. Specifically, we observe that by applying our \dynamvit{} to DeiT-256, we obtain a model that has a comparable computational complexity to DeiT-Ti, but enjoys around $4.3\%$ higher ImageNet top-1 accuracy. }

\newpart{To further illustrate the acceleration ability of our method, we compare the throughput of our dynamic spatial sparsification method with model scaling on GPU in Fig.~\hyperref[fig:tradeoff]{5(b)}. The throughput is measured on a single NVIDIA RTX 3090 GPU with batch size fixed to 128. For our \dynamvit{} with DeiT-S as the base model, we test the throughput on different $\rho$ varying from 1.0 to 0.5. For comparisons, we test the throughput of DeiT models with decreasing embedding dimensions and keep the similar computation complexity to our models. We can observe that our \dynamvit{} model run much faster than width scaling models with similar FLOPs, which clearly shows the potential of our model to accelerate the inference process in more practical settings. }

\newpart{\paragrapha{Dynamic spatial sparsification V.S. channel pruning.} Conventional model acceleration methods usually perform pruning on channel dimensions~\cite{he2017channel}, which has been thoroughly studied in previous work. Our dynamic spatial sparsification method discovers the potential of acceleration in a brand new direction by taking advantage of the asymmetry in spatial dimensions. Thus we suppose that these two kinds of methods can be applied to model acceleration jointly to fully consider the redundancy on different dimensions and boost performance. To illustrate this, we use the recent state-of-the-art vision Transformer channel reduction method AutoFormer~\cite{chen2021AutoFormer} as our baseline and apply our method to it. AutoFormer~\cite{chen2021AutoFormer} is an efficient neural architecture search (NAS) network by training supernet and searching for the optimal subnets as the efficient models, which can be viewed as a strong baseline of channel pruning~\cite{liu2018rethinking}. Firstly, we search for the best architecture from supernet and pick the best-performance ones in pre-defined complexity ranges. As the picked subnet is already well-trained, we can directly train our \dynamvit{} based on them without further tuning. The results are plotted in Fig.~\hyperref[fig:tradeoff]{5(c)}. We can observe that the model can be further compressed in spatial dimensions after touching the limit of channel pruning methods. }

\begin{figure*}
    \centering
    \includegraphics[width=\textwidth]{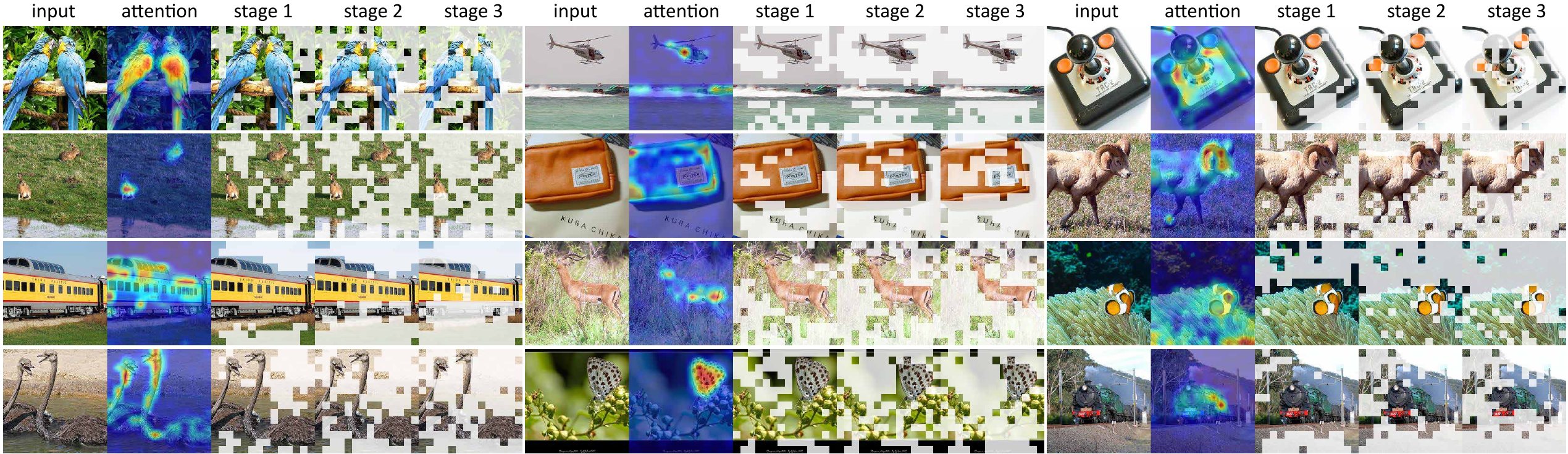}\vspace{-5pt}
    \caption{\small \textbf{Visualization of the progressively sparsified tokens on ImageNet.}  We show the original input image, attention visualizations of transformer models using method~\cite{chefer2020transformer}, and the sparsification results after the three stages, where the masks represent the corresponding tokens that are discarded. We see our method can gradually focus on the most representative regions in the image, especially activated regions of attention maps, and has better interpretability. }
    \label{fig:viz}
\end{figure*}

\paragrapha{Comparisons of different sparsification strategies.} As illustrated in Fig.~\ref{fig:idea}, the dynamic spatial sparsification produces unstructured features. To discuss whether dynamic sparsification is better than other strategies, we perform ablation experiments and the results are shown in Table~\ref{tab:ablation}. For the structural downsampling, we perform an average pooling with kernel size $2\times 2$ after the sixth block of the baseline DeiT-S~\cite{touvron2020deit} model, which has similar FLOPs to our DynamicViT. Static token sparsification means that the sparsification decisions are not conditioned on the input tokens. We also compare our method with other token removal methods like randomly removing tokens or removing tokens based on the attention score of the class token. We find through the experiments that although other strategies have similar computational complexities, the proposed dynamic token sparsification method achieves the best accuracy. We also show that the progressive sparsification method is significantly better than one-stage sparsification.

\newpart{For hierarchical models, we investigate the effects of different fast path designs for dynamic spatial sparsification in Table~\ref{tab:strategy}. We tried linear projection layer with a \texttt{Linear}($C$, $C$) block and a bottleneck MLP architecture with a \texttt{Linear}($C$, $C/4$) $\to$ \texttt{GELU} $\to$ \texttt{Linear}($C/4$, $C$) block. We can observe from the results that though bottleneck architecture can lead to relatively smaller complexity, it will also slightly hurt the performance. We also tried replacing the previous network with simpler designs such as learnable masks and zero masks, where we directly replace uninformative features with learnable parameters and zeros respectively. The results suggest that such designs will significantly hurt performance. We conjecture the input-independent fast paths may break the neighboring relations learned from the pre-training process and have poorer expressive power.}

\begin{table}[t]
  \centering
    \caption{\small \newpart{\textbf{Effects of different fast path designs for dynamic spatial sparsification for hierarchical models.} We investigate four different strategies to replace the feed-forward networks based on DynamicCNN-S/0.9. We can observe that the simple linear projection achieves the best trade-off. The bottleneck MLP architecture has a slight effect on performance with fewer FLOPs while filling uninformative features with learnable parameters or zeros will significantly hurt the performance. } }\vspace{-5pt}
    \adjustbox{width=\linewidth}{
    \begin{tabu}to 0.9\linewidth{lXX}\toprule
    Strategy & \multicolumn{1}{l}{Top-1 Acc. (\%)}&\multicolumn{1}{l}{FLOPs (G)}\\\midrule
    Feed-Forward Network & 83.1 & 8.70\\\midrule
    Linear Projection &83.1\cb{(-0.0)}&6.78\cb{(-22\%)}\\
    Bottleneck MLP &82.9\cb{(-0.2)}&6.63\cb{(-24\%)}\\
    Learnable Mask &55.5\cb{(-26.6)}&6.48\cb{(-26\%)}\\
    Zero Mask &53.2\cb{(-29.9)}&6.48\cb{(-26\%)}\\\bottomrule
    \end{tabu}%
    }
  \label{tab:strategy}%
\end{table}%

\begin{table}[t]
  \centering
    \caption{\small \newpart{\textbf{Ablations on different losses.} We provide the results after removing the distillation loss and the KL loss based on DynamicViT-S/0.7. We can observe that the distill loss $\mathcal{L}_{\rm distill}$ and KL loss $\mathcal{L}_{\rm KL}$ both contribute to better performance. } }\vspace{-5pt}
    \adjustbox{width=\linewidth}{
    \begin{tabu}to 0.9\linewidth{lXX}\toprule
    Base Model&\multicolumn{1}{l}{DeiT-S}&\multicolumn{1}{l}{LVViT-S}\\\midrule
    DynamicViT&79.3\cb{(-0.5)}&83.0\cb{(-0.3)}\\
    w/o distill (Eq.13)&79.3\cb{(-0.5)}&82.7\cb{(-0.6)}\\
    w/o KL (Eq.14)&79.2\cb{(-0.6)}&82.9\cb{(-0.4)}\\
    w/o distill \& KL&79.2\cb{(-0.6)}&82.5\cb{(-0.8)}\\\bottomrule
    \end{tabu}%
    }
  \label{tab:loss}%
\end{table}%

\begin{table}[t]
  \centering
  \caption{\small \newpart{\textbf{Ablations on keeping ratio and pruning location for hierarchical models.} We conduct experiments based on DynamicCNN-T. The model has [3,3,9,3] layers in four stages  ($S^1-S^4)$. Keeping ratio $\rho=1.0$ signifies no sparsification is applied. For fair comparisons, we fix the complexity of the accelerated models to 3.6G FLOPs. }}\vspace{-5pt}
  \adjustbox{width=\linewidth}{
  \setlength{\tabcolsep}{10pt}
    \begin{tabular}{cccccc}\toprule
    \multirow{2}{*}{$S^1$} & \multirow{2}{*}{$S^2$} & \multicolumn{2}{c}{$S^3$} & \multirow{2}{*}{$S^4$} & \multirow{2}{*}{Top-1 Acc. (\%)}\\
    \cmidrule(lr){3-4}
    & & Ratio $\rho$ & Loc & \\\midrule
     {\cg{1.0}} & {\cg{1.0}} & 0.7 & [1,2,3] & {\cg{1.0}} & 81.6{\cb{(-0.5)}} \\
     {\cg{1.0}} & {\cg{1.0}} & 0.7 & [0,1,2] & {\cg{1.0}} & 81.2{\cb{(-0.9)}} \\
     {\cg{1.0}} & {\cg{1.0}} & 0.6 & [2,4,6] & {\cg{1.0}} & 80.9{\cb{(-1.2)}} \\\midrule
    0.7 & 0.7 & 0.8 & [1,2,3] & {\cg{1.0}} & 81.5{\cb{(-0.6)}} \\
    {\cg{1.0}} & 0.7 & 0.8 & [1,2,3] & 0.7 & 79.7{\cb{(-2.4)}} \\
    0.7 & {\cg{1.0}} & 0.8 & [1,2,3] & 0.7 & 78.9{\cb{(-3.2)}} \\\midrule
    0.5 & {\cg{1.0}} & 0.8 & [1,2,3] & {\cg{1.0}} & 81.5{\cb{(-0.6)}} \\
    {\cg{1.0}} & 0.5 & 0.8 & [1,2,3] & {\cg{1.0}} & 81.5{\cb{(-0.6)}} \\
    {\cg{1.0}} & {\cg{1.0}} & 0.8 & [1,2,3] & 0.5 & 80.3{\cb{(-1.8)}} \\
    \bottomrule
    \end{tabular}%
    }
  \label{tab:layer}%
\end{table}%

\begin{table}[t]
\caption{\small \textbf{Results on the higher input resolution.} We apply our method with a larger input size (\ie, DeiT-S with $384\times 384$ input) and compare the image classification accuracy with normally used $224\times 224$ input size at keeping ratio $\rho=0.7/0.5$.} \label{tab:large}\vspace{-5pt}
\centering
\adjustbox{width=\linewidth}{
\begin{tabular}{lllll}
    \toprule
        \multirow{2}{*}{Model} & \multicolumn{2}{c}{$224\times 224$} & \multicolumn{2}{c}{$384\times 384$} \\
        \cmidrule(lr){2-3}\cmidrule(lr){4-5}
        & FLOPs (G) & Acc. (\%) & FLOPs (G) & Acc. (\%) \\ \midrule
        DeiT-S & 4.6 & 79.8 & 15.5 & 81.6 \\ 
        DyViT-S/0.7 & 3.0\cb{(-35\%)} & 79.3 \cb{(-0.5)} & 9.5 \cb{(-39\%)} & 81.4 \cb{(-0.2)} \\
        DyViT-S/0.5 & 2.2\cb{(-52\%)} & 77.5 \cb{(-2.3)} & 7.0 \cb{(-55\%)} & 80.3 \cb{(-1.3)} \\ \bottomrule
    \end{tabular}
}
\end{table}

\paragrapha{Visualizations.} To further investigate the behavior of our proposed dynamic spatial sparsification method, we visualize the sparsification procedure of \dynamvit{} in Fig.~\ref{fig:viz}. We show the original input image and the sparsification results after the three stages, where the masks represent the corresponding tokens that are discarded. We find that through the hierarchically token sparsification, our \dynamvit{} can gradually drop the uninformative tokens and finally focus on the objects in the images. This phenomenon also suggests that the~\dynamvit{} leads to better interpretability, \ie, it can locate the important parts in the image which contribute most to the classification step-by-step.

\begin{figure*}
    \centering
    \includegraphics[width=\textwidth]{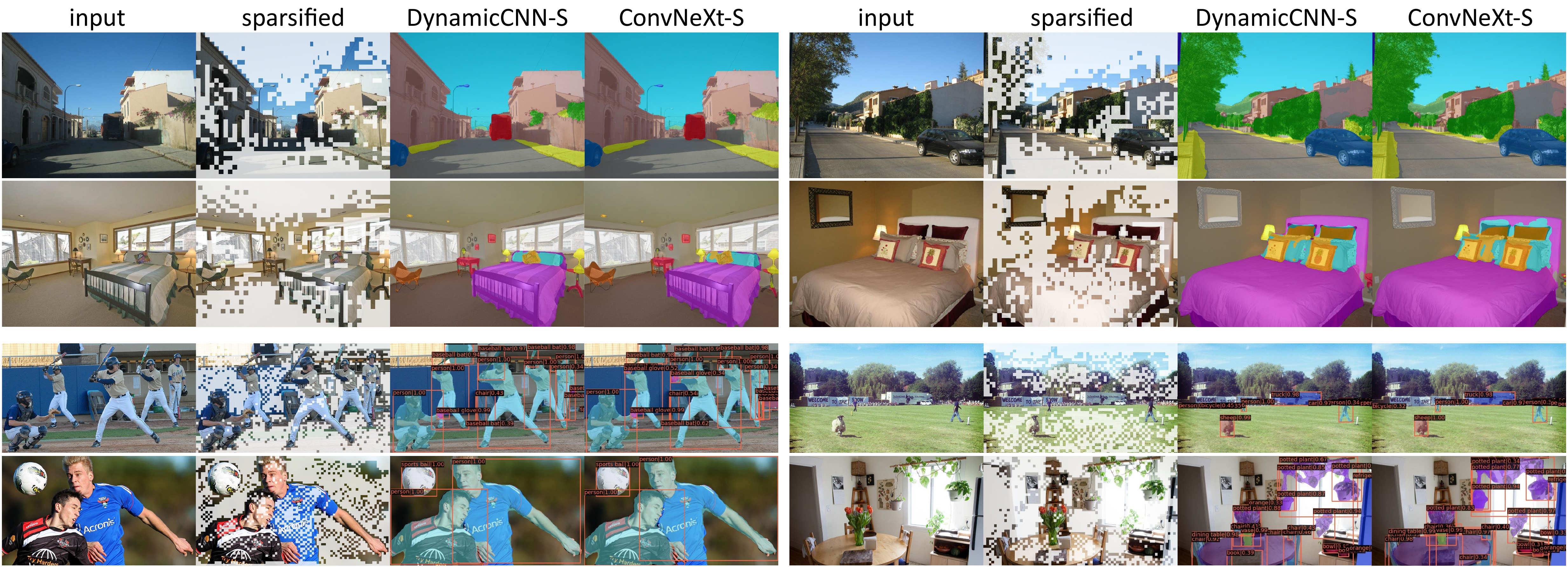}\vspace{-5pt}
    \caption{\small \newpart{\textbf{Visualization of the sparsification and prediction results on downstream tasks.} The top two rows are from the semantic segmentation task while the bottom two rows are from the object detection task. We use ConvNeXt-S as the baseline model and use our DynamicCNN-S/0.9 model to prune 50\% of the regions. We present the final sparsified images after three sparsification stages and compare the results of downstream tasks by offering the predictions after and before pruning. Visualization results suggest that our method can focus on informative objects with limited influence on the predictions.}}
    \label{fig:downstream}
\end{figure*}

\newpart{To illustrate how our dynamic spatial sparsification work on downstream tasks with larger images and multiple objects, we visualize the sparsification and prediction results on ADE20k and COCO validation set in Fig.~\ref{fig:downstream}. We show the input images and sparsification results after pruning 50$\%$ of the regions (\ie $\rho=0.9$) through our lightweight prediction modules, together with the final prediction of the models. We find that our DynamicCNN can recognize the uninformative parts of the images (\eg background or interior of the objects) and keep the key objects and boundaries remaining, thus our method can reduce the complexity while consistently maintaining the performance on dense prediction tasks. }

\begin{figure}[t]
\centering
\centering
\includegraphics[width=0.48\textwidth]{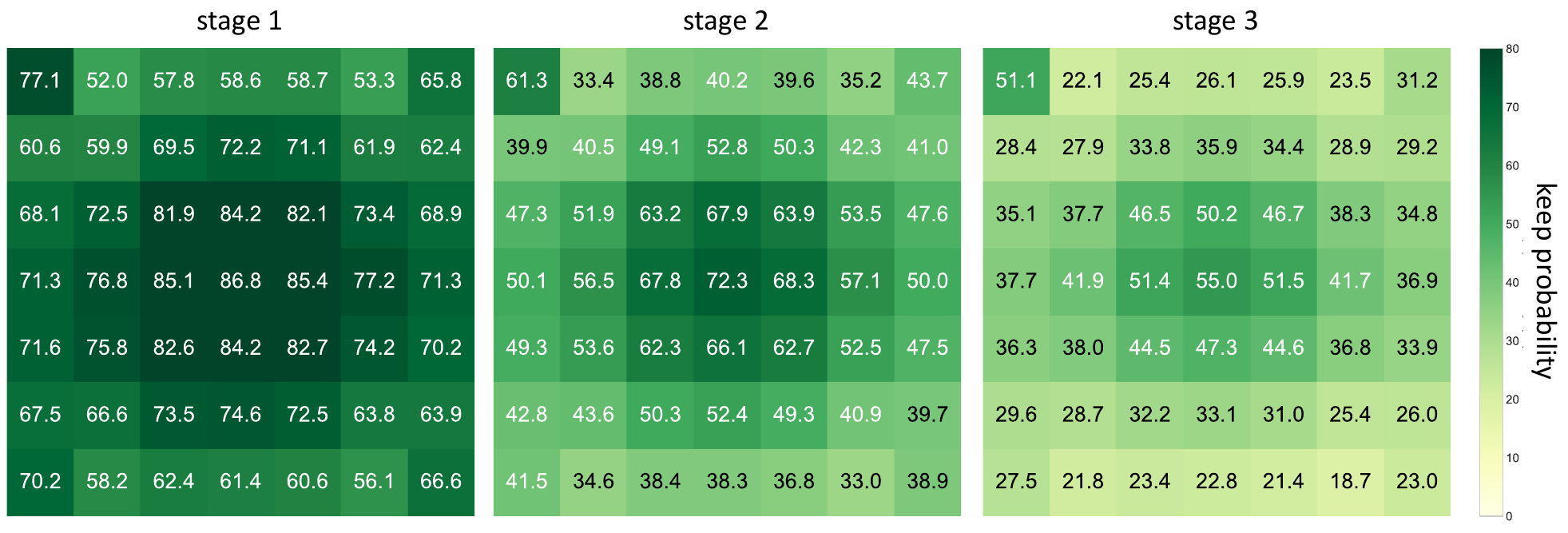}\vspace{-5pt}
\caption{\small \textbf{Keeping probabilities of the tokens at each stage.} We use our DynamicViT-S/0.7 model to generate the decisions for all the images on the ImageNet validation set and compute the keeping probability of each token in all three stages. }\label{fig:keep_probl} 
\end{figure}

Besides the sample-wise visualization we have shown above, we are also interested in the statistical characteristics of the sparsification decisions, \ie, what kind of general patterns does our dynamic spatial sparsification method learn from the dataset? We then use the \dynamvit{} to generate the decisions for all the images in the ImageNet validation set and compute the keeping probability of each token in all three stages, as shown in Fig.~\ref{fig:keep_probl}. We average pool the probability maps into $7\times 7$ such that they can be visualized more easily. Unsurprisingly, we find the tokens in the middle of the image tend to be kept, which is reasonable because in most images the objects are located in the center. We can also find that the later stage generally has lower probabilities to be kept, mainly because the keeping ratio at the $s$ stage is $\rho^s$, which decreases exponentially as $s$ increases.

\paragrapha{Ablations on different losses.} We show the effects of different losses in Table~\ref{tab:loss}. Although the improvement brought by the distillation loss and the KL loss is not very significant, it can consistently boost the performance of various models.

\newpart{\paragrapha{Ablations on keeping ratio and pruning location for hierarchical models.} As hierarchical models are more complex in design than isotropic vision Transformers (\eg ViTs), we conduct ablation studies on the position of applying our lightweight prediction module. As stage 3 usually has significantly more layers than other stages in most hierarchical models, we focus on two key questions: which layers in stage 3 and which stage to apply sparsification. The results are shown in Table~\ref{tab:layer}. We conduct our experiments by fixing the target computation complexity to 3.6G FLOPs for fair comparisons. We can observe from the results that applying sparsification at the first three layers can achieve better performance than applying sparsification evenly while leaving the first layer after downsampling unchanged can also boost performance. We can observe that applying sparsification to all four stages will also hurt the overall performance of the model and simply performing sparsification at stage 3 can lead to the best performance. According to the results, we take the best pruning location (\ie stage 3 and layer [$k,2k,3k$] with $k=\lfloor \rm Layer/9\rfloor$) in all experiments and adjust $k$ accordingly for larger models with more layers. }

\newpart{\paragrapha{Accelerating models with higher resolutions.} To show the effectiveness of our method on higher resolutions, we apply our method to the larger input size (\ie, DeiT-S with $384\times 384$ input). The results are presented in Table~\ref{tab:large}. We see our method also works well with the larger input. The accuracy drop becomes less significant when we apply our method to the model with larger feature maps. Notably, we can reduce the complexity of the DeiT-S model with $384\times 384$ input by over 50\% with only a 1.3\% accuracy drop. We can observe that the drop of accuracy is relatively lower compared with $224\times 224$ input size at the same keeping ratio, which means that a larger pruning ratio can be applied to the input with more tokens and our method can achieve a better trade-off with the larger input size. }

\begin{figure}[t]
\centering
\includegraphics[width=0.48\textwidth]{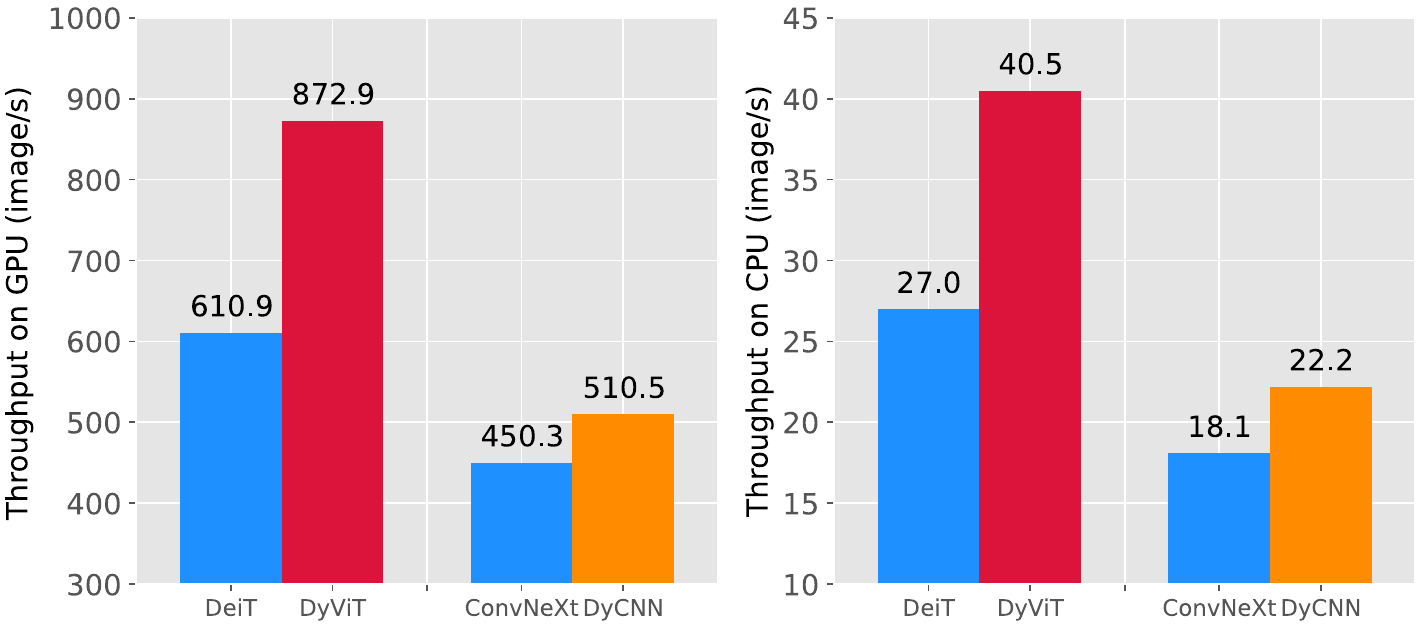}\vspace{-5pt}
\caption{\small \textbf{Comparisons of throughput on GPU and CPU}. We compare the throughput of our dynamic spatial sparsification method with vision Transformer (DeiT-B) and CNN (ConvNeXt-B) baselines. We can observe that our method can accelerate the inference process on multiple devices and architectures.}
\label{fig:throughput}
\end{figure}

\newpart{\paragrapha{Comparisons on throughput on GPU and CPU. } We test the throughput of baseline models and our dynamic spatial sparsification models on both GPU and CPU. The throughput is measured on a single NVIDIA RTX 3090 GPU and Intel Xeon Gold 6321C CPU with batch size fixed to 128. For fair comparisons, we keep the same  FLOPs for different models and fix the complexity reduction ratio of our models at 34\% (\ie keeping ratio $\rho=0.7$). Our \dynamvit{} can increase the throughput by over 40\% on both GPU and CPU while our DynamicCNN can increase the throughput by 13\% on GPU and 23\% on CPU. It should be noted that the throughput of DeiT-B is much higher than ConvNeXt-B, which shows that the hierarchical models run slower on hardware while having higher classification accuracy on ImageNet (\ie 83.8\% for ConvNeXt-B v.s. 81.8\% for DeiT-B). The actual acceleration rate of \dynamvit{} is relatively higher than DynamicCNN, which suggests that the token sparsification framework is more hardware-friendly. Since our split and reassemble operations are not fully optimized, we believe the throughput of our hierarchical models can be further improved with better implementation.  }

\section{Conclusion}
\label{sec:con}
\newpart{In this paper, we have studied accelerating vision models by exploiting spatial sparsity. We have proposed a dynamic spatial sparsification framework that can be applied to various prevalent backbones including isotropic vision Transformers, modern CNNs and hierarchical vision Transformers. We have evaluated our framework on both image classification and dense prediction tasks. Our results demonstrated that our method can effectively reduce the computational cost during inference for various backbones and tasks with negligible performance drops. We hope that our attempt can inspire future work to further explore spatial sparsity in vision tasks. }

\section*{Acknowledgement}

This work was supported in part by the National Key
Research and Development Program of China under Grant
2017YFA0700802, in part by the National Natural Science
Foundation of China under Grant 62125603, Grant
61822603, Grant U1813218, Grant U1713214, in part by
Beijing Academy of Artificial Intelligence (BAAI), and in
part by a grant from the Institute for Guo Qiang, Tsinghua
University.

\vspace{-5pt}

\appendix
\section*{A. Implementation Details}\label{sec:details}

\paragrapha[0pt]{Details about image classification experiments.} We conduct our experiments on the ImageNet (also known as ILSVRC2012)~\cite{deng2009imagenet} dataset. ImageNet is a commonly used benchmark for image classification. We train our models on the training set, which consists of 1.28M images. The top-1 accuracy is measured on the 50k validation images following common practice~\cite{he2016deep,touvron2020deit}. To fairly compare with previous methods, we report the single crop results. 

We fix the number of sparsification stages $S=3$ in all of our experiments, since this setting can lead to a decent trade-off between complexity and performance.  For the prediction module, we use the identical architecture for different stages. We use two \texttt{LayerNorm} $\to$ \texttt{Linear}($C$, $C/2$) $\to$ \texttt{GELU} block to produce $\mathbf{z}^{\rm local}$ and $ \mathbf{z}^{\rm global}$  respectively. We employ a \texttt{Linear}($C$, $C/2$) $\to$ \texttt{GELU} $\to$ \texttt{Linear}($C/2$, $C/4$) $\to$ \texttt{GELU} $\to$ \texttt{Linear}($C/4$, $2$) $\to$ \texttt{Softmax} block to predict the probabilities.

During training our DynamicViT models, we follow most of the training techniques used in DeiT~\cite{touvron2020deit}. We use the pre-trained vision Transformer models to initialize the backbone models and jointly train the backbone model as well as the prediction modules for 30 epochs with a cosine learning rate schedule and 10-epoch linear warm-up. We set the learning rate of the prediction modules to $\frac{\text{batch size}}{1024}\times 0.001$ and use $0.01\times$  learning rate for the backbone model. The batch size is adjusted adaptively for different models according to the GPU memory.  We fix the weights of the backbone models in the first 5 epochs. 

During training our DynamicCNN and DynamicSwin models, we follow most of the above-mentioned training techniques. Different from our experiments of DynamicViT, we train the models for 120 epochs since we need to optimize the new fast path layers. We use the EMA model~\cite{polyak1992acceleration} to avoid over-fitting and set the drop path ratio to 0.2, 0.2 and 0.5 for tiny, small and base models respectively. We set the learning rate of the backbone model as  $0.2\times$ of the prediction modules. 

\paragrapha{Details about semantic segmentation experiments.} We use the popular semantic FPN framework~\cite{kirillov2019panoptic} in our semantic segmentation experiments on ADE20k~\cite{zhou2017scene}. We use the AdamW~\cite{loshchilov2017decoupled} optimizer and train the model for 40k iterations with a batch size of 32. We set the initial learning rate and weight decay to 0.0001 and 0.05 respectively. We train our model and the baseline using the identical training configuration for fair comparisons.  We use a pre-trained semantic segmentation with the same architecture as the teacher model to compute the distillation loss. We set the drop path ratio to 0.3 and 0.4 for small and base models respectively.

\paragrapha{Details about object detection experiments.} We use the widely used Mask-RCNN~\cite{he2017mask} framework to conduct the object detection and instance segmentation experiments on COCO~\cite{lin2014coco}. We follow the $1\times$ schedule to the models for 12 epochs with a batch size of 16. We use the AdamW~\cite{loshchilov2017decoupled} optimizer to train our models and baseline models. We use the same initial learning rate and weight decay as our segmentation experiments, and adopt a similar strategy to compute the distillation loss. We set the drop path ratio to 0.6 and 0.7 for small and base models respectively.

\section*{B. More Analysis}


In this section, we provide more analysis of our method. We investigate the effects of progressive sparsification, distillation loss, ratio loss, and keeping ratio. We also include more visualization results. The following describes the details of the experiments, results and analysis. 

\paragrapha{Analysis of progressive sparsification. } To verify the effectiveness of the progressive sparsification strategy, we test different sparsification methods that result in similar overall complexity. Here we provide more detailed results and more analysis. We find that progressive sparsification is much better than single-shot sparsification. Increasing the number of stages will lead to better performance. Since further increasing the number of stages ($>3$) will not lead to significantly better performance but add computation, we use a 3-stage progressive sparsification strategy in our main experiments. 

\begin{table}[!h]
    \centering \small
    \newcolumntype{g}{>{\columncolor{Gray}}l}
   \begin{tabular}{gll}\toprule
     & \multicolumn{1}{l}{Top-1 acc. (\%)} & \multicolumn{1}{l}{GFLOPs} \\\midrule
    DeiT-S~\cite{touvron2020deit} & 79.8  & 4.6 \\\midrule
    $\rho = 0.25, [\rho]$ (single-stage) & 77.4\cb{(-2.4)} & 2.9\cb{(-37\%)} \\
    $\rho = 0.60, [\rho, \rho^2]$ (two-stage) & 79.2\cb{(-0.6)}  & 2.9\cb{(-37\%)} \\
    $\rho = 0.70, [\rho, \rho^2, \rho^3]$ (three-stage) & 79.3\cb{(-0.5)}  & 2.9\cb{(-37\%)} \\\bottomrule
\end{tabular}
\end{table}

\paragrapha{Ablation on the distillation loss and ratio loss. } The weights of the distillation losses and ratio loss are the key hyper-parameters in our method.  Since the token-wise distillation loss and the KL divergence loss play similar roles in our method, we set $ \lambda_{\rm KL} = \lambda_{\rm distill}$ in all of our experiments for the sake of simplicity. In this experiment, we fix the keeping ratio $\rho$ to be 0.7.  We find our method is not sensitive to these hyper-parameters in general. The proposed ratio loss can encourage the model to reach the desired acceleration rate.  Distillation losses can improve the performance after sparsification.  We directly apply the best hyper-parameters searched on DeiT-S for all models.

\begin{table}[!h]
    \centering \small
    \setlength{\tabcolsep}{9pt}
    \newcolumntype{g}{>{\columncolor{Gray}}l}
\begin{tabular}{ggl}\toprule
     & & \multicolumn{1}{l}{Top-1 accuracy (\%)} \\\midrule
    DeiT-S~\cite{touvron2020deit} & & 79.8  \\\midrule
    $ \lambda_{\rm KL} = \lambda_{\rm distill} = 0$ & {\color{gray} $\lambda_{\rm ratio}= 2$ }& 79.17\cb{(-0.63)}  \\
    $ \lambda_{\rm KL} = \lambda_{\rm distill} = 0.5$ & {\color{gray}  $\lambda_{\rm ratio}= 2$}& 79.32\cb{(-0.48)} \\
    $ \lambda_{\rm KL} = \lambda_{\rm distill} = 1 $ & {\color{gray} $\lambda_{\rm ratio}= 2$} & 79.23\cb{(-0.57)} \\\midrule
    {\color{gray} $\lambda_{\rm KL} = \lambda_{\rm distill} = 0.5$} & $\lambda_{\rm ratio}= 1$ & 79.15\cb{(-0.65)}  \\
    {\color{gray} $\lambda_{\rm KL} = \lambda_{\rm distill} = 0.5$} & $ \lambda_{\rm ratio}= 2$& 79.32\cb{(-0.48)} \\
    {\color{gray} $\lambda_{\rm KL} = \lambda_{\rm distill} = 0.5$} & $ \lambda_{\rm ratio}= 4$ & 79.29\cb{(-0.51)} \\
    \bottomrule
\end{tabular}
\end{table}

\paragrapha{Smaller keeping ratio. } We have also tried applying a smaller keeping ratio (larger acceleration rate). The results based on DeiT-S~\cite{touvron2020deit} and LV-ViT-S~\cite{jiang2021token} models are presented in the following tables. We see that using $\rho < 0.7$ will lead to a significant accuracy drop while reducing fewer FLOPs. Since only 22\% and 13\% tokens are remaining in the last stage when we set $\rho$ to 0.6 and 0.5 respectively, small $\rho$ may cause a significant information loss. Therefore, we use $\rho \geq 0.7$ in our main experiments. Jointly scaling  $\rho$ and the model width can be a better solution to achieve a large acceleration rate as shown in Figure 5 in the paper.

\begin{table}[!h]
    \centering \small
\newcolumntype{g}{>{\columncolor{Gray}}l}
  \begin{tabular}{gllll}\toprule
    \multirow{2}{*}{} & \multicolumn{2}{c}{DeiT-S)} & \multicolumn{2}{c}{LVViT-S}\\
    \cmidrule(lr){2-3}  \cmidrule(lr){4-5} & Top-1 (\%) & GFLOPs & Top-1 (\%) & GFLOPs \\\midrule
    
    Baseline & 79.8  & 4.6  & 83.3  & 6.6  \\\midrule
    $\rho = 0.9$  & 79.8\cb{(-0.0)} & 4.0\cb{(-14\%)}   & 83.3\cb{(-0.0)} & 5.8\cb{(-12\%)} \\
    $\rho = 0.8$ & 79.6\cb{(-0.3)}  & 3.4\cb{(-27\%)} & 83.2\cb{(-0.1)}  & 5.1\cb{(-22\%)} \\
     $\rho = 0.7$  & 79.3\cb{(-0.5)} & 2.9\cb{(-37\%)}  & 83.0\cb{(-0.3)} & 4.6\cb{(-31\%)} \\
    $\rho = 0.6$ & 78.5\cb{(-1.3)}  & 2.5\cb{(-46\%)} & 82.6\cb{(-0.7)}  & 4.1\cb{(-38\%)} \\
    $\rho = 0.5$ & 77.5\cb{(-2.3)}  & 2.2\cb{(-52\%)}  & 82.0\cb{(-1.3)}  & 3.7\cb{(-44\%)}  \\\bottomrule
\end{tabular}
\end{table}

\paragrapha{Comparisons with the baselines with longer training. } Since our method needs to train the base model for additional 30 and 120 epochs for DynamicViT and DynamicCNN/DynamicSwin respectively. We provide the performance of our base models with longer training. We can see longer training will not significantly improve the performance of our baseline models, which suggests the small accuracy drops mainly benefit from our framework instead of longer training.
\begin{table}[!h]
    \centering \small
\newcolumntype{g}{>{\columncolor{Gray}}l}
  \begin{tabular}{glll}\toprule
   Model & Training length & Top-1 (\%) & GFLOPs \\ \midrule
    DeiT-S~\cite{touvron2020deit} & 300 epochs & 79.8 & 4.6\\
    DeiT-S~\cite{touvron2020deit} & 330 epochs & 79.9\cb{(+0.1)} & 4.6\\
    DyViT-S/0.7 & 300+30  epochs & 79.3\cb{(-0.5)} & 2.9\cb{(-37\%)} \\ \midrule
    ConvNeXt-T~\cite{liu2022convnet} & 300 epochs & 82.1 & 4.5 \\
    ConvNeXt-T~\cite{liu2022convnet} & 420 epochs & 82.1\cb{(+0.0)} & 4.5 \\
    DyCNN-T/0.7 & 300+120 epochs & 81.6\cb{(-0.5)} & 3.6\cb{(-20\%)} \\
     \bottomrule
\end{tabular}
\end{table}

\section*{C. More Visual Results}

\begin{figure*}
    \centering
    \includegraphics[width=\textwidth]{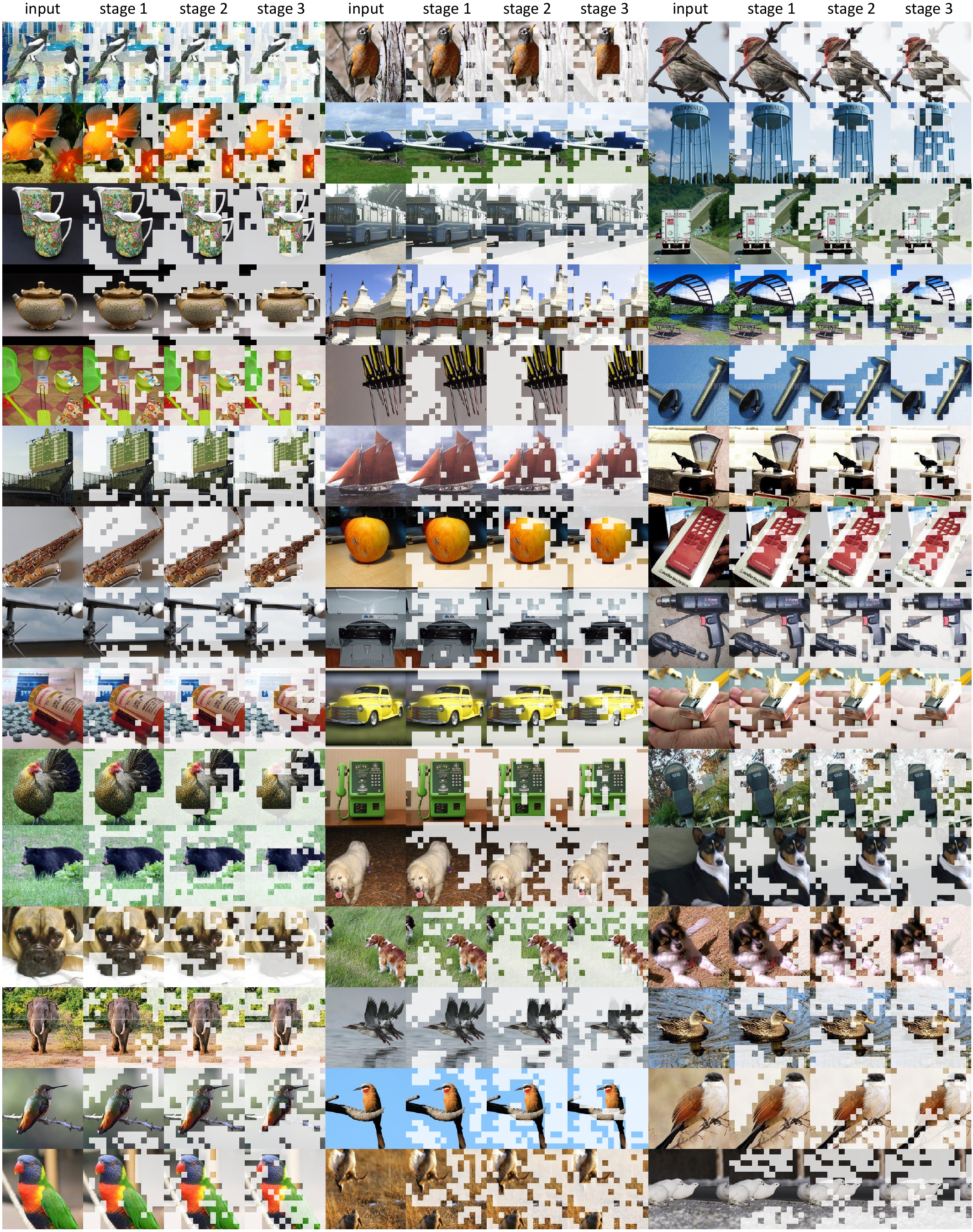}
    \caption{\textbf{More visual results of DynamciViT}. The input images are randomly sampled from the validation set of ImageNet. }
    \label{fig:viz1}
\end{figure*}

\begin{figure*}
    \centering
    \includegraphics[width=\textwidth]{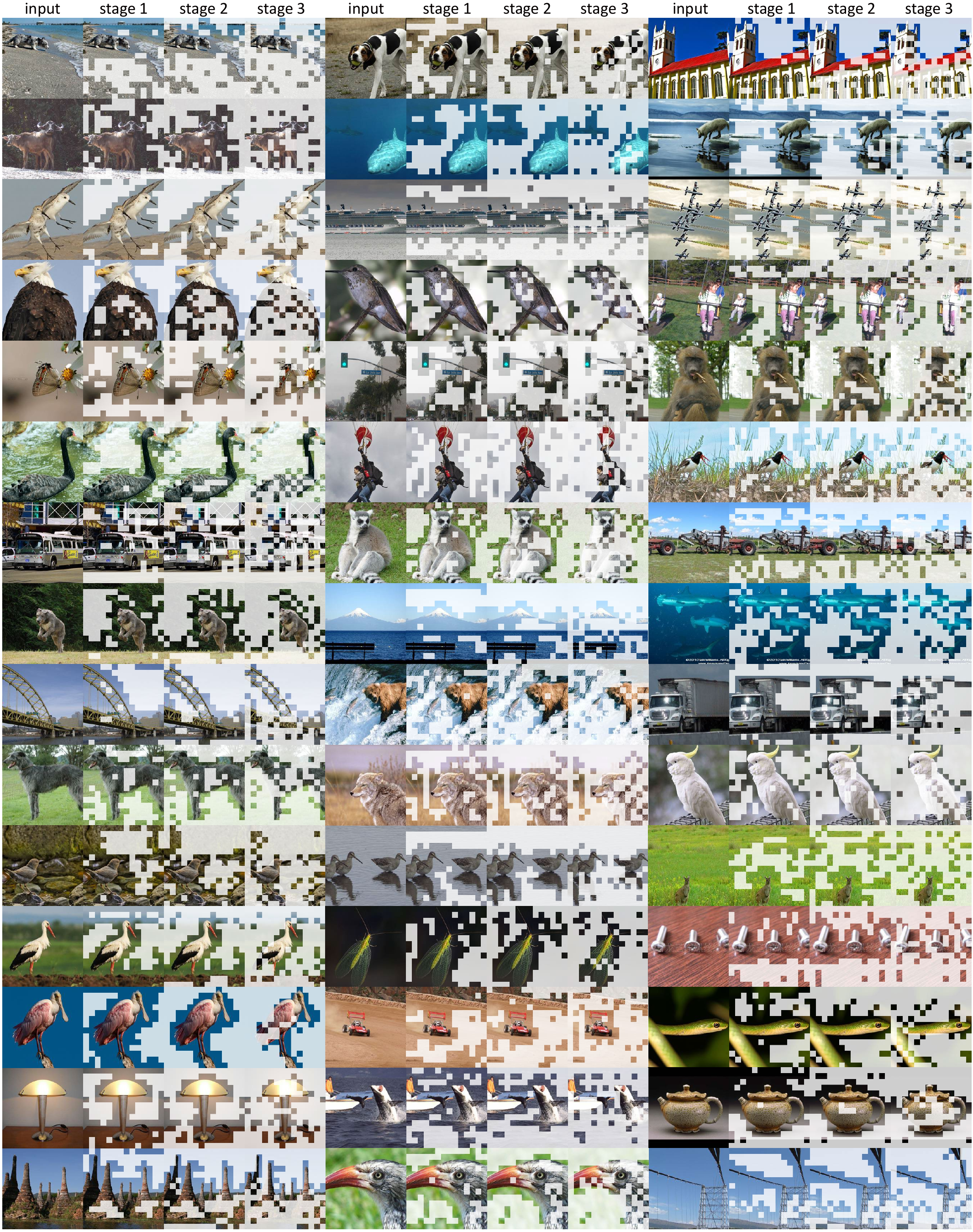}
    \caption{\textbf{More visual results of DynamicCNN}. The input images are randomly sampled from the validation set of ImageNet. }
    \label{fig:viz2}
\end{figure*}

\begin{figure*}
    \centering
    \includegraphics[width=\textwidth]{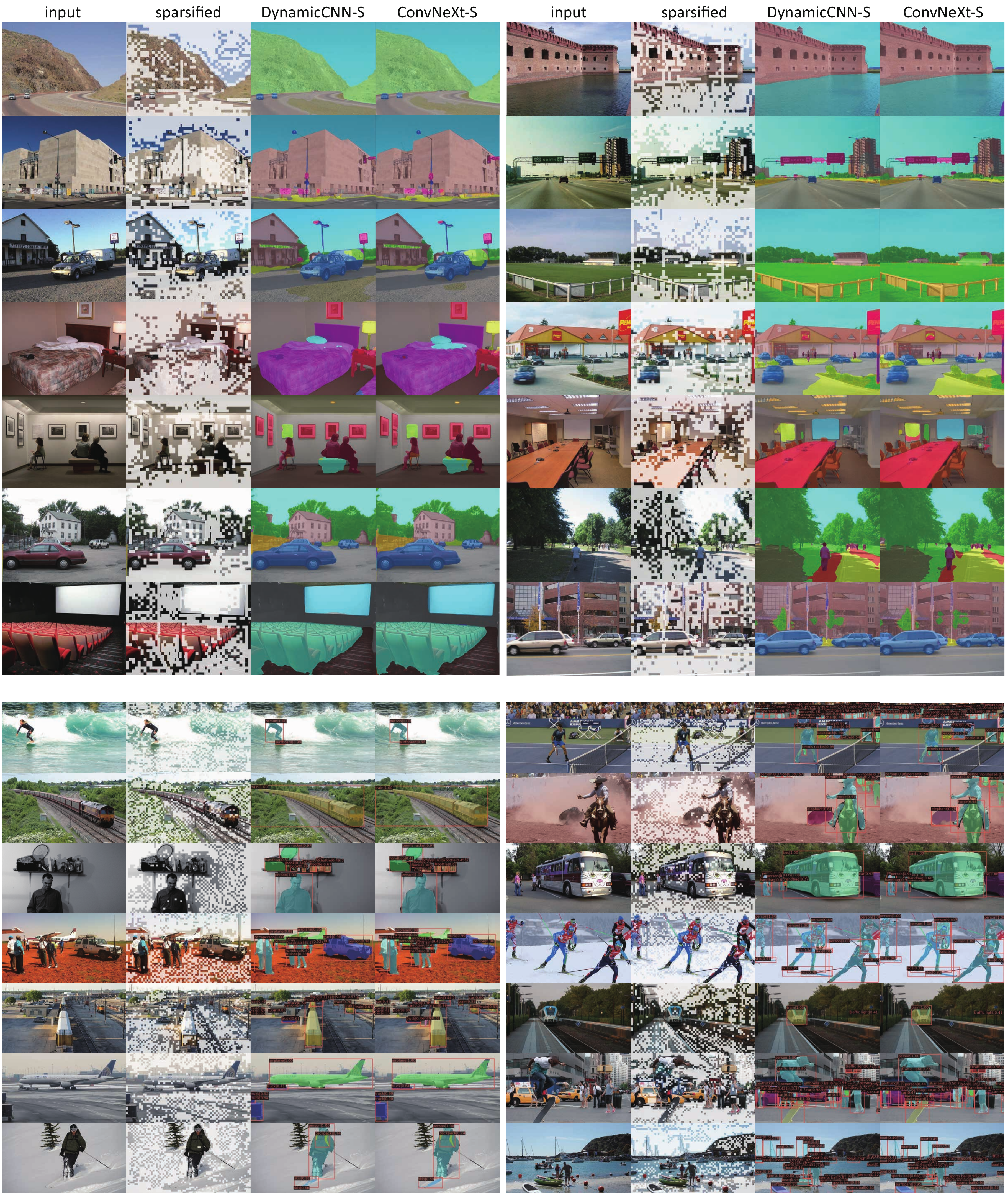}
    \caption{\textbf{More visual results of our models for semantic segmentation and object detection}. The input images are randomly sampled from the validation set of ADE20k and COCO.}
    \label{fig:viz3}
\end{figure*}

We provide more visual results of our classification models based on vision Transformers (Fig.~\ref{fig:viz1}) and hierarchical models  (Fig.~\ref{fig:viz2}), and our dense prediction models (Fig.~\ref{fig:viz3}). We see that the models equipped with our dynamic spatial sparsification framework can gradually focus on the most informative regions and progressively filter out the redundant patches. Although we introduce the new fast path layers for hierarchical models, we can observe similar sparsification patterns in these two types of models. 

\bibliographystyle{IEEEtranS}
\bibliography{ref}

\end{document}